\documentclass{article} % For LaTeX2e
\usepackage{iclr2027_conference,times}

\usepackage{amsmath,amsfonts,bm}

\def\eqref#1{equation~\ref{#1}}
\def\1{\bm{1}}

\DeclareMathAlphabet{\mathsfit}{\encodingdefault}{\sfdefault}{m}{sl}
\SetMathAlphabet{\mathsfit}{bold}{\encodingdefault}{\sfdefault}{bx}{n}

\usepackage{hyperref}
\usepackage{url}
\usepackage{graphicx}
\usepackage{subcaption}
\usepackage{booktabs}
\usepackage{multirow}
\usepackage{booktabs}
\usepackage{makecell}
\usepackage[table]{xcolor}
\usepackage{enumitem}
\usepackage{algorithm}
\usepackage{algorithmic}
\usepackage{bbm}

\title{Routing Should Pay for Itself: Sparse \\ Supervision for Economical LLM Routing}

\author{%
\vspace{6pt}%
\normalfont\mdseries%
Guannan Lai$^{1,2,4}$ \quad
Gelin Bian$^{1,2}$ \quad
Hao-Xuan Ma$^{1,2}$ \quad
Jun-Peng Jiang$^{1,2}$%
\\[2pt]
Long Chen$^{3}$ \quad
Jian-Dong Liu$^{4}$ \quad
Zhi-Hao Tan$^{1,2}$ \quad
Han-Jia Ye$^{1,2}$%
\\[5pt]
{\small
$^{1}$School of Artificial Intelligence, Nanjing University%
}\\[-1pt]
{\small
$^{2}$National Key Laboratory for Novel Software Technology, Nanjing University%
}\\[-1pt]
{\small
$^{3}$The Hong Kong University of Science and Technology
\qquad
$^{4}$SinapisAI%
}%
\\[4pt]
{\footnotesize
\texttt{\{laign, biangl, mahx, jiangjp, liujd, tanzh, yehj\}@lamda.nju.edu.cn}%
}\\[-1pt]
{\footnotesize
\texttt{longchen@ust.hk}%
}%
}

\iclrfinalcopy % Uncomment for camera-ready version, but NOT for submission.
\begin{document}

\maketitle
\lhead{Preprint}

\begin{abstract}
Large language model (LLM) routing reduces serving cost by assigning each query
to an appropriate model while preserving response quality.
Learning such a router, however, often requires executing multiple candidate
models on historical queries to collect query--model quality feedback, creating
a nontrivial supervision cost before deployment.
Existing work largely focuses on serving-time efficiency, overlooking whether
the resulting savings are sufficient to recover this upfront expenditure.
We further observe that routing quality often saturates well before all
query--model feedback is collected, suggesting that dense supervision can be
economically over-provisioned.
We propose \textsc{SaveRouter}, a sparse-supervision routing framework that
selectively acquires informative model feedback and shares capability
information across related queries, while retaining query-level refinement for
fine-grained routing.
We evaluate routing by jointly accounting for supervision expenditure and
subsequent serving-time savings.
Across four routing benchmarks, the main setting uses only about 33--41\% of
available training feedback while maintaining competitive or better routing
quality, and reduces the break-even deployment volume by approximately
1.9--9.5$\times$ compared with the fastest conventional router.
Further analysis shows that acquiring more supervision is not always
economically preferable: the supervision level that minimizes serving cost can
differ from the one that achieves the earliest payback.
Our code is publicly available at
\url{https://github.com/LAMDA-Model-Reuse/SaveRouter}.
\end{abstract}

\section{Introduction}

Large language models (LLMs) differ substantially in both capability and
serving cost~\citep{chen2024frugalgpt,vsakota2024fly}, creating opportunities
to reduce inference expense by routing each query to an appropriate model
rather than always invoking the strongest one.
This has motivated a growing body of LLM routing methods that optimize the
quality--cost trade-off at serving time
~\citep{ding2024hybrid,ong2024routellm,zhuang2025embedllm}.
Yet existing evaluations typically begin only after the router has been
constructed, focusing on how much it saves during deployment.
In practice, many supervised routers must first execute candidate models on
historical queries to collect query--model quality feedback
~\citep{zhuang2025embedllm,mei2025omnirouter,shi2025inference}.
This feedback acquisition incurs an upfront supervision expenditure before
any serving-time savings can be realized.

\begin{figure*}[t]
    \centering

    \begin{subfigure}[t]{0.48\textwidth}
        \centering
        \includegraphics[width=\linewidth]{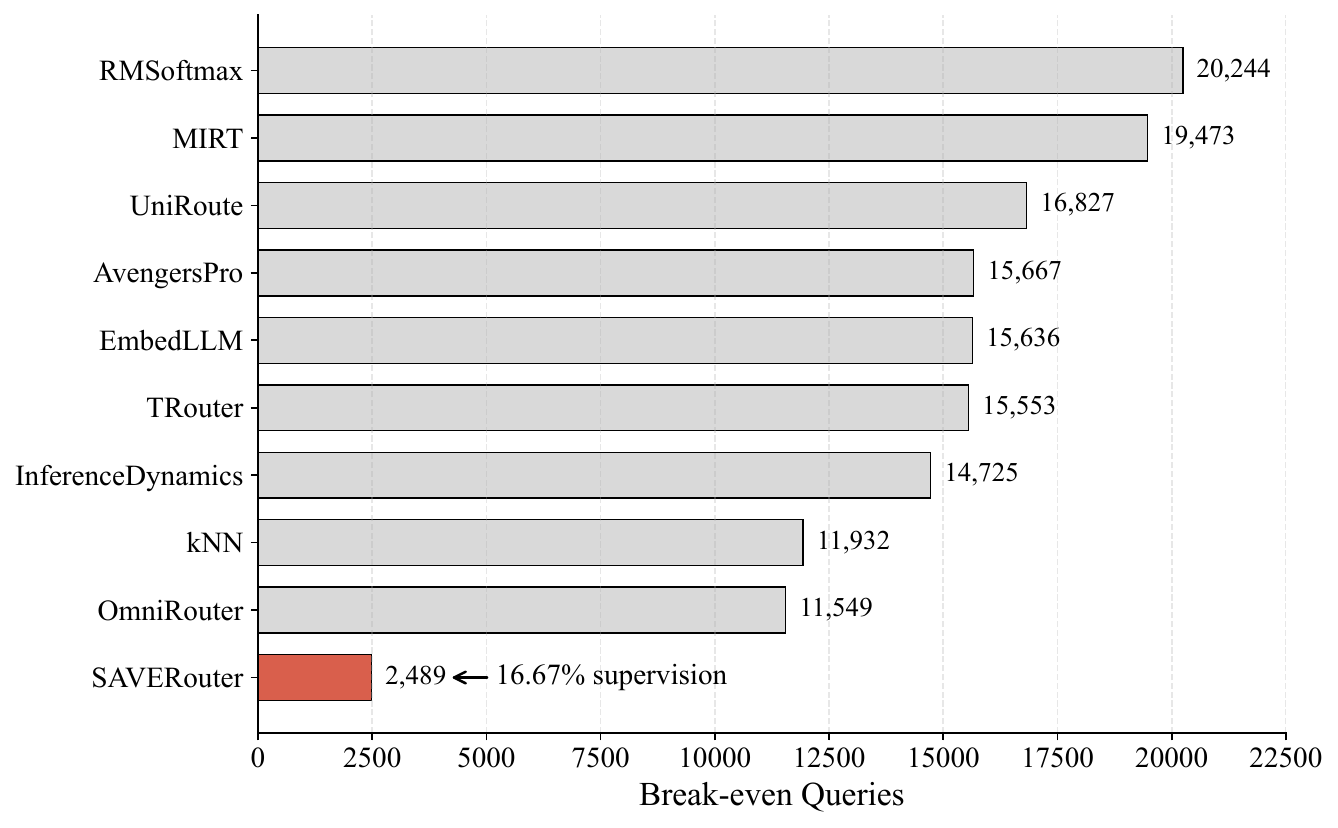}
        \caption{Upfront supervision delays routing payback.}
        \label{fig:payback}
    \end{subfigure}
    \hfill
    \begin{subfigure}[t]{0.48\textwidth}
        \centering
        \includegraphics[width=\linewidth]{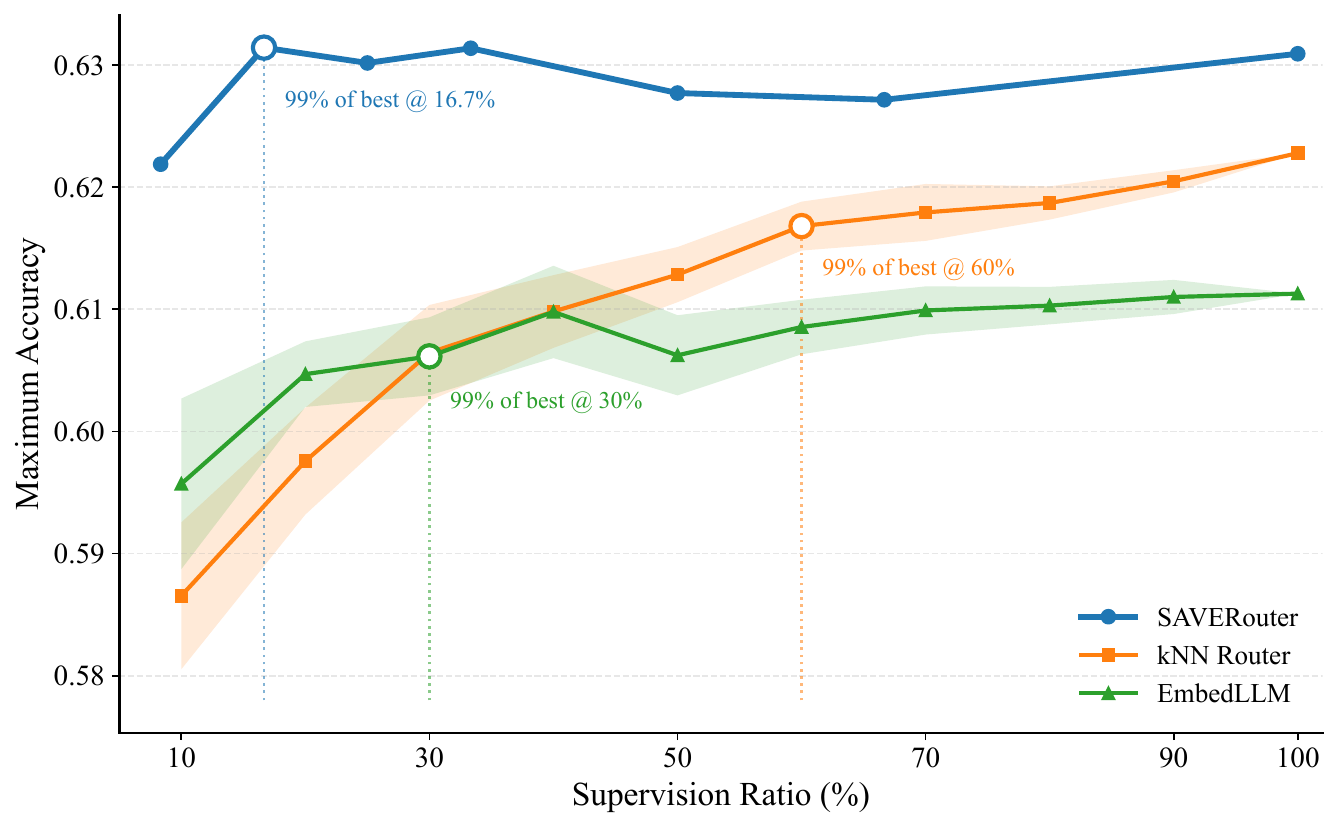}
        \caption{Routing quality saturates before full supervision.}
        \label{fig:sparse_supervision}
    \end{subfigure}

    \caption{
    \textbf{Dense supervision can be economically over-provisioned.}
    Left: supervision expenditure can substantially delay the break-even point
    of routers with lower serving-time cost.
    Right: routing quality often saturates well before the full query--model
    matrix is observed; EmbedLLM and kNN recover 99\% of their fully supervised
    accuracy with only 30\% and 60\% supervision, respectively.
    }
    \vspace{-0.5cm}\label{fig:intro_motivation}
\end{figure*}

This supervision expenditure can be substantial.
For $N$ training queries and $M$ candidate models, densely evaluating the
query--model matrix requires up to $N\times M$ model executions before
deployment.
Consequently, lower serving cost after deployment does not immediately imply
an economic gain: the accumulated serving-time savings must first offset the
cost of acquiring routing supervision.
As illustrated in Figure~\ref{fig:intro_motivation}(a), this upfront
expenditure can substantially delay the point at which a router pays for
itself.
At the same time, dense supervision may provide far more feedback than routing
actually needs.
Figure~\ref{fig:intro_motivation}(b) shows that EmbedLLM and kNN reach 99\% of
their full-supervision accuracy with only 30\% and 60\% supervision,
respectively.
This gap between supervision expenditure and marginal routing benefit suggests
that dense query--model evaluation can be economically over-provisioned.

These observations motivate a sparse-supervision setting in which only a
small fraction of query--model outcomes can be acquired before deployment.
This setting introduces two coupled challenges.
First, under a limited supervision budget, we must determine which
query--model outcomes are most informative for downstream routing, rather
than spending evaluations uniformly.
Second, the resulting feedback is sparse and non-uniform, requiring the router
to infer unobserved model behavior from limited evidence while retaining
query-specific variation.
We therefore ask:
\emph{how can we acquire only the feedback needed for effective routing and
learn reliably from it, so that the upfront supervision expenditure is
recovered as early as possible?}

To address this problem, we propose \textsc{SaveRouter}, a sparse-supervision
routing framework designed to extract more routing-relevant information from
each acquired feedback signal.
Rather than evaluating every model on every training query, \textsc{SaveRouter}
adaptively selects a small set of informative query--model outcomes and exploits
shared capability structure across related queries to estimate the unobserved
ones.
A query-level correction further captures fine-grained variation within each
group, producing model-quality estimates for cost-aware routing.
By reducing the supervision expenditure required before deployment while
preserving effective routing decisions, \textsc{SaveRouter} allows the upfront
investment in routing supervision to be recovered much earlier.

Serving-time quality and cost alone do not reveal whether the savings from
routing are sufficient to recover the supervision expenditure incurred before
deployment.
We therefore introduce two complementary metrics: \textbf{SA-BEP} measures how
many deployment queries are needed for serving-time savings to recover the
upfront supervision expenditure, while \textbf{SA-CR} measures the resulting
cost ratio after amortizing this expenditure over a fixed deployment horizon.
Across four heterogeneous routing benchmarks, \textsc{SaveRouter} uses only
about 33--41\% of the available training feedback while maintaining competitive
or better routing quality.
At the target quality, it reduces the break-even deployment volume by
approximately 1.9--9.5$\times$ compared with the fastest conventional
fully supervised router.
Further analysis shows that routing quality often saturates well before full
supervision is collected, and that additional supervision does not necessarily
lead to earlier payback or lower amortized cost.

In summary, our contributions are as follows:
\begin{itemize}[leftmargin=*]
    \item We account for upfront supervision expenditure in LLM routing and
    introduce \textbf{SA-BEP} and \textbf{SA-CR} to quantify payback and amortized cost.

    \item We propose \textsc{SaveRouter}, which learns effective routing from sparse
    feedback through adaptive acquisition and structured capability estimation.

    \item Experiments on four benchmarks show that \textsc{SaveRouter} maintains
    competitive routing quality with much less supervision and earlier break-even.
\end{itemize}

\section{Related Work}

\subsection{LLM Routing and Model Selection}

LLM routing selects an appropriate model for each query to balance response
quality and inference cost.
One line of work adopts \emph{cascading}, where a cheaper model is queried
first and its response is used to determine whether a stronger model is needed.
FrugalGPT and AutoMix follow this paradigm through response scoring and
self-verification, respectively
\citep{chen2024frugalgpt,aggarwal2024automix}.
Predictive routing instead selects a model before response generation.
Some methods learn when a cheaper model is sufficient:
Hybrid LLM predicts whether a query should be escalated to a stronger model,
while RouteLLM learns routing preferences from comparison data
\citep{ding2024hybrid,ong2024routellm}.
Other methods explicitly model the relationship between queries and model
capabilities.
RouterDC learns query--model representations through contrastive learning,
EmbedLLM learns compact model representations, and GraphRouter captures
relationships among tasks, queries, and models using a heterogeneous graph
\citep{chen2024routerdc,zhuang2025embedllm,feng2025graphrouter}.
While these approaches differ in how routing decisions are modeled, they
typically assume access to substantial query--model supervision during router
construction.

\subsection{Routing with Sparse or Partial Supervision}

Recent work relaxes the dense-supervision assumption by learning from limited
or partial feedback.
SemiRouter uses data-rich anchor models and a lightweight adapter to incorporate
new models from sparse training data \citep{wang2026semirouter}.
BaRP learns routing policies from bandit feedback, where only the outcome of
the selected model is observed, while allowing the quality--cost preference to
vary at inference time \citep{wei2025learning}.
WISERouter jointly considers exploration and routing under a workload-level
budget that covers both data collection and deployment
\citep{li2026wiserouter}.

These works show that effective routing can be learned without dense feedback,
but they target different supervision settings and objectives.
Our focus is the upfront supervision expenditure of router construction: given
a limited supervision budget, which query--model outcomes should be acquired,
and how quickly can serving-time savings recover this investment?
\textsc{SaveRouter} addresses this setting through adaptive sparse feedback
acquisition and structured capability estimation, and evaluates routing in
terms of both serving efficiency and break-even behavior.
\section{Routing Should Pay for Itself}
\label{sec:problem}

Existing LLM routing methods primarily evaluate the quality--cost trade-off
after router construction, overlooking the supervision cost incurred before
deployment.
We therefore account for this upfront expenditure when evaluating routing,
asking whether serving-time savings can eventually recover the supervision
investment.

\subsection{The Upfront Cost of Learning to Route}
\label{sec:upfront}

Consider a training set
$\mathcal{X}=\{x_i\}_{i=1}^{N}$ and a candidate model pool
$\mathcal{M}=\{1,\ldots,M\}$.
Let $y_{im}$ denote the quality of model $m$ on query $x_i$.
Obtaining $y_{im}$ requires executing and evaluating the corresponding model.
For an observed set of query--model pairs
$\Omega\subseteq[N]\times\mathcal{M}$, we define the upfront supervision cost as
\[
    C_0(\Omega)
    =
    \sum_{(i,m)\in\Omega} a_{im},
    \label{eq:supervision_cost}
\]
where $a_{im}$ is the acquisition cost of observing $y_{im}$.
Under dense supervision,
$\Omega_{\mathrm{dense}}=[N]\times\mathcal{M}$,
requiring $NM$ model evaluations.
Thus, router construction becomes increasingly expensive as either the
training workload or the candidate pool grows.

This upfront expenditure directly affects whether routing is economically
useful.
Let $C_{\mathrm{ref}}$ denote the average serving cost of a reference system,
and let $C_{\mathrm{serve}}(\pi)$ be the average serving cost of routing policy
$\pi$ under the same quality requirement.
Its per-query saving is
\(
    \Delta C(\pi)
    =
    C_{\mathrm{ref}}-C_{\mathrm{serve}}(\pi).
\)
After serving $T$ deployment queries, the cumulative net saving is
\(
    S(T;\pi,\Omega)
    =
    T\,\Delta C(\pi)-C_0(\Omega).
    \label{eq:net_saving}
\)
When $\Delta C(\pi)>0$, the supervision investment is recovered after
approximately
\begin{equation}
    \mathrm{SA\mbox{-}BEP}
=
\frac{C_0(\Omega)}{\Delta C(\pi)}.
    \label{eq:break_even}
\end{equation}

Equation~\ref{eq:break_even} exposes a limitation of evaluating routers only
by their serving-time cost.
Two routers with similar serving-time efficiency can have very different
payback behavior if one requires substantially more supervision to construct.
Reducing supervision cost therefore shortens the time needed to recover the
upfront investment.
This expression makes the trade-off explicit: reducing supervision is useful
only insofar as the resulting router preserves sufficient serving-time savings.
The objective is therefore not to minimize $C_0$ in isolation, but to reduce
upfront supervision without sacrificing the quality--cost advantage that
eventually amortizes it.

\subsection{Dense Supervision Is Over-Provisioned}
\label{sec:redundancy}

Dense supervision provides complete query--model feedback, but completeness
is stronger than what routing actually requires.
We identify two sources of redundancy.

\noindent \textbf{Structured redundancy.}
Related queries often exhibit similar model-performance patterns, allowing
feedback on one query to inform model capability on others.
Treating query--model pairs independently therefore ignores shared structure
in the supervision matrix.

\noindent \textbf{Decision redundancy.}
More importantly, routing is a decision problem rather than a matrix-recovery
problem.
For a cost preference $\lambda$, the router ultimately needs to identify
\[
    m^{*}(x)
    =
    \arg\max_{m\in\mathcal{M}}
    \left[
        y_m(x)-\lambda c_m(x)
    \right],
    \label{eq:routing_decision}
\]
rather than accurately estimate every $y_m(x)$.
Additional observations that refine capability estimates without changing the
selected model provide little value to the final routing decision.

Dense supervision therefore optimizes \emph{information completeness},
whereas routing only requires \emph{decision sufficiency}.
As illustrated in Figure~\ref{fig:intro_motivation}, routing quality can remain competitive
under substantially reduced supervision, while the lower acquisition cost
directly shortens the break-even horizon.
This suggests that supervision should itself be treated as a scarce resource:
the important question is not whether every outcome can be observed, but which outcomes are most informative and decision-relevant for effective routing.

\subsection{Sparse-Supervision LLM Routing}
\label{sec:sparse_setting}

Motivated by this observation, we consider a setting in which only a small
fraction of query--model outcomes can be acquired during router construction.
Let $O_{im}\in\{0,1\}$
indicate whether $y_{im}$ is observed, and define
\[
    \Omega
    =
    \{(i,m)\mid O_{im}=1\}.
\]
Given a supervision budget of at most $K\ll M$ models per training query,
\[
    \sum_{m=1}^{M} O_{im}\le K,
    \qquad \forall i.
    \label{eq:sparse_budget}
\]
The router is trained only from
\[
\mathcal{D}_{\mathrm{sparse}}
=
\{(x_i,m,y_{im},c_{im})\mid(i,m)\in\Omega\},
\]
where selecting a query--model pair reveals both its quality feedback
$y_{im}$ and realized serving cost $c_{im}$.
All remaining query--model outcomes are unavailable during router
construction.

Under this setting, supervision acquisition and router learning become
coupled.
Formally, we seek an observation set $\Omega$ and the resulting routing policy
$\pi_{\Omega}$ that maximize deployment utility under a limited supervision
budget:
\[
    \begin{aligned}
    \max_{\Omega,\pi_{\Omega}}
    \quad &
    \mathbb{E}_{x\sim\mathcal{D}}
    \left[
        y_{\pi_{\Omega}(x)}(x)
        -
        \lambda c_{\pi_{\Omega}(x)}(x)
    \right]
    \\
    \mathrm{s.t.}\quad&
    |\Omega_i|\le K,
    \qquad \forall i,
\end{aligned}
\label{eq:sparse_objective}
\]
where
$\Omega_i=\{m\mid(i,m)\in\Omega\}$.

The goal is therefore \textbf{not} to reconstruct the complete query--model matrix.
Instead, sparse-supervision routing seeks the smallest amount of informative
feedback sufficient to preserve effective routing decisions.
This gives rise to two coupled challenges:
\textit{feedback acquisition}, which determines which query–model outcomes are most informative for downstream routing decisions, and
\emph{sparse capability estimation}, which infers model behavior from the
limited observations.
\textbf{\textsc{SaveRouter} } addresses these two challenges by adaptively acquiring informative
feedback and exploiting shared capability structure across related queries,
as described next.

\section{\textsc{SaveRouter} : Learning to Route from Sparse Supervision}
\label{sec:method}

\begin{figure}[t]
    \centering
    \includegraphics[width=\linewidth]{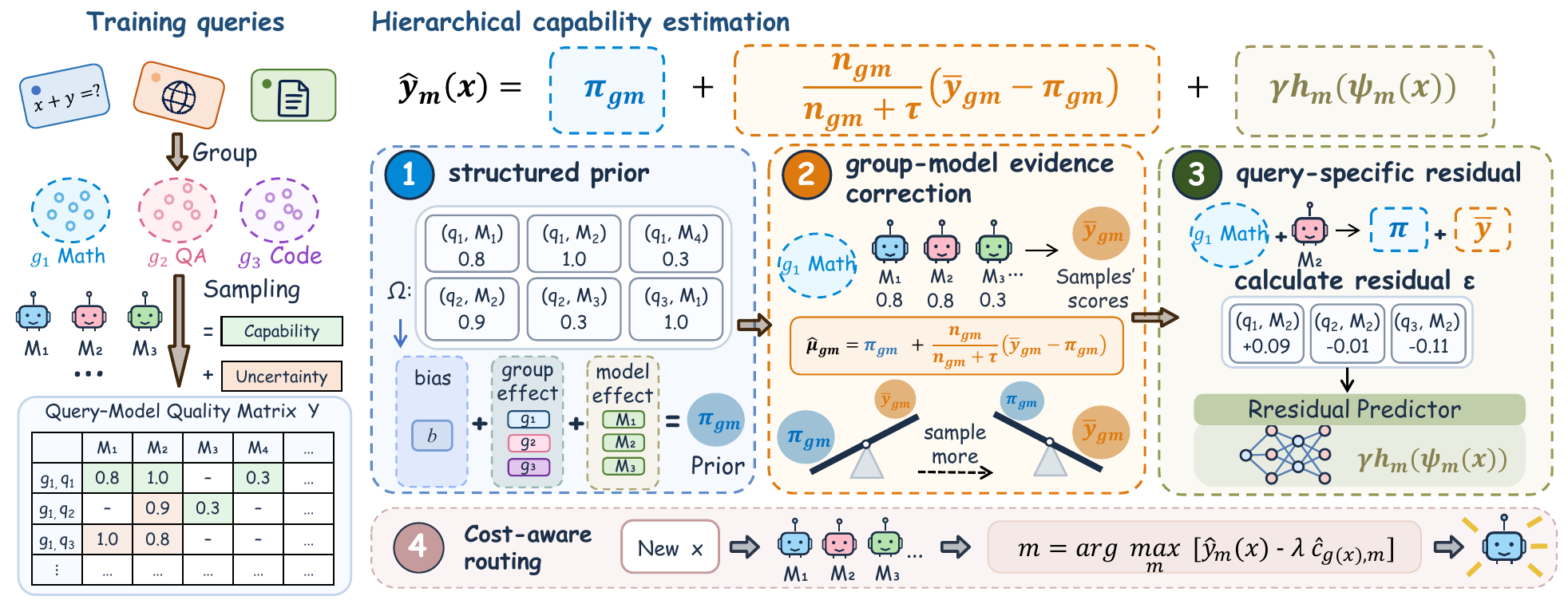}
    \caption{
    \textbf{Overview of \textsc{SaveRouter} .}
    \textsc{SaveRouter}  first groups related training queries and adaptively acquires
    sparse query--model feedback using capability and uncertainty.
    It then estimates model capability through a structured group--model prior,
    evidence-based correction from the acquired observations, and a
    query-specific residual predictor.
    At deployment time, the resulting quality estimates are combined with
    estimated serving costs for cost-aware routing.
    }
    \label{fig:SaveRouter_overview}
\end{figure}

\textsc{SaveRouter}  is designed around the two challenges identified in
Section~\ref{sec:problem}: deciding which feedback is most informative to
acquire and inferring model capability from sparse observations.
Figure~\ref{fig:SaveRouter_overview} provides an overview of the framework.
The overall procedure consists of three stages.
First, \textsc{SaveRouter}  groups related training queries and adaptively allocates a
small evaluation budget using both estimated capability and uncertainty,
yielding a sparse set of observed query--model outcomes.
Second, it performs hierarchical capability estimation: a structured
group--model prior shares information across groups and models, acquired
evidence corrects this prior, and a lightweight residual predictor recovers
query-specific variation.
Finally, the resulting quality estimates are combined with serving-cost
estimates to perform cost-aware routing for unseen queries.

\subsection{Adaptive Sparse Feedback Acquisition}
\label{sec:acquisition}

With only $K\ll M$ model evaluations available per query, uniformly sampling
candidate models may waste supervision on models whose behavior is already
well understood or unlikely to affect routing decisions.
\textsc{SaveRouter}  instead allocates feedback according to both estimated capability
and uncertainty.

\noindent\textbf{Query grouping.}
The acquisition process relies on the observation that related queries often
share similar model-performance patterns.
We therefore assign each query $x_i$ to a group $g_i$.
When predictable task labels are available in the training data, they define
the training groups and we learn a classifier to assign unseen queries;
otherwise, we cluster frozen query representations.
Grouping uses only query inputs and training-side group information, never
model-quality feedback.

\noindent \textbf{Group-conditioned capability tracking.}
For each group--model pair $(g,m)$, we maintain its observation count
$n_{gm}$ and cumulative quality $S_{gm}$.
Because some pairs may receive very few observations, their empirical means
can be unstable.
We therefore first estimate the global capability of model $m$ and then
shrink the group-specific estimate toward it:
\[
    \bar{\mu}_m
    =
    \frac{S_m+\alpha_0}
         {n_m+\alpha_0+\beta_0},
    \qquad
    \tilde{\mu}_{gm}
    =
    \frac{S_{gm}+\tau_0\bar{\mu}_m}
         {n_{gm}+\tau_0}.
    \label{eq:acq_estimate}
\]
Here, $\bar{\mu}_m$ captures the overall capability of model $m$, while
$\tilde{\mu}_{gm}$ adapts this estimate to group $g$ as group-specific
evidence accumulates.

\noindent \textbf{Capability--uncertainty acquisition.}
Acquisition proceeds for $K$ passes over the training set, with one new
query--model outcome acquired per query in each pass.
At the beginning of each pass, we reset the group- and model-level acquisition
statistics and randomly permute all training queries, while retaining the
observation mask across passes.
Using only feedback revealed earlier in the current pass, we score each
group--model pair by
\begin{equation}
    a_{gm}
    =
    \tilde{\mu}_{gm}
    +
    \beta_{\mathrm{ucb}}
    \sqrt{
        \frac{
        \log\!\left(\sum_j n_{gj}+1\right)}
        {\max(n_{gm},1)}
    },
    \qquad
    m_i
    =
    \arg\max_{m\in\mathcal{C}_i}
    a_{g_i m},
    \label{eq:acquisition}
\end{equation}
where $\mathcal{C}_i$ contains models that are available for $x_i$ and have not
been previously acquired for this query.
If the current group contains candidate models with $n_{g_i m}=0$, we
uniformly select one such model before applying Eq.~(2), ensuring basic
within-pass coverage.
After selecting $m_i$, we reveal its feedback and update the current-pass
statistics.
Each pass therefore contributes one new observation per training query, and
after $K$ passes the sparse supervision contains $NK$ distinct
pairs.

\subsection{Hierarchical Capability Estimation}
\label{sec:hierarchical}

Adaptive acquisition produces sparse and highly non-uniform observations.
Consequently, directly using the empirical mean of each group--model pair is
unreliable: frequently observed pairs may be estimated accurately, whereas
rare or completely unobserved pairs provide little or no local evidence.
Moreover, observations collected under the adaptive policy are not uniformly
distributed across groups and models.
\textsc{SaveRouter}  therefore estimates capability hierarchically, combining shared
global structure with local group-level evidence.
After acquisition is complete, we aggregate observations across all $K$
passes.
We use $S^{\Omega}_{gm}$ and $n^{\Omega}_{gm}$ to denote the cumulative
quality sum and observation count for group--model pair $(g,m)$ over the
complete sparse observation set $\Omega$.

\noindent \textbf{Structured group--model prior.}
We first fit a two-way additive model on the observed pairs:
\[
    y_{im}
    =
    b + u_{g_i} + v_m + \epsilon_{im},
    \qquad
    (\hat b,\hat u,\hat v)
    =
    \arg\min_{b,u,v}
    \sum_{(i,m)\in\Omega}
    (y_{im}-b-u_{g_i}-v_m)^2
    +
    \lambda_{\mathrm{prior}}
    \left(\|u\|_2^2+\|v\|_2^2\right).
    \label{eq:prior}
\]
Here, $b$ represents overall task difficulty, $u_g$ captures systematic
differences among query groups, and $v_m$ captures global differences among
candidate models.
Because these parameters are shared across many observations, the model
provides a stable estimate even for sparsely observed group--model pairs.
The resulting structured prior is
\[
    \pi_{gm}
    =
    \operatorname{clip}
    \left(
        \hat b+\hat u_g+\hat v_m,\,
        0,1
    \right).
    \label{eq:group_prior}
\]

\noindent \textbf{Local evidence with shrinkage.}
The prior captures shared structure but cannot replace direct observations
when sufficient local evidence is available.
We therefore combine it with the observed outcomes for each group--model pair:
\[    
    \hat{\mu}_{gm}
=
\frac{S^{\Omega}_{gm}+\tau\pi_{gm}}
     {n^{\Omega}_{gm}+\tau}
=
\frac{n^{\Omega}_{gm}}
     {n^{\Omega}_{gm}+\tau}\bar y_{gm}
+
\frac{\tau}
     {n^{\Omega}_{gm}+\tau}\pi_{gm}.
    \label{eq:hierarchical_estimate}
\]
The parameter $\tau$ controls the strength of the prior.
When $n_{gm}$ is large, the estimate is dominated by observed group-specific
feedback; when observations are scarce, it increasingly relies on shared
structure.
For an entirely unobserved pair, the estimate naturally reduces to the
structured prior.

\subsection{Query-Level Residual Correction}
\label{sec:residual}

The hierarchical estimate in Eq.~\ref{eq:hierarchical_estimate} assigns the
same group-level capability to all queries within a group.
This provides a stable estimate under sparse supervision, but inevitably
misses within-group variation.
For example, two queries assigned to the same task group may still differ in
difficulty or favor different models.

We therefore learn only the variation that remains unexplained by the shared
group structure.
For each observed pair $(i,m)$, we remove its direct contribution from the
local group--model statistics and construct
\[
\hat{\mu}^{(-i)}_{g_i m}
=
\frac{
S^{\Omega}_{g_i m}-y_{im}+\tau\pi_{g_i m}
}{
n^{\Omega}_{g_i m}-1+\tau
},
\qquad
e_{im}
=
y_{im}-\hat{\mu}^{(-i)}_{g_i m}.
\]
Removing $y_{im}$ from the local sufficient statistics reduces direct
self-influence through the group-level estimate.
For each model $m$ with sufficient acquired observations, we fit a lightweight predictor
\[
h_m
=
\arg\min_h
\sum_{i:(i,m)\in\Omega}
\left(
e_{im}-h(\psi_{\mathrm{res}}(x_i))
\right)^2
+
\lambda_{\mathrm{ctx}}\|h\|_2^2,
\]
where $\psi_{\mathrm{res}}(x)$ denotes a lightweight contextual representation
that is separate from the representation used for query grouping.
The final capability estimate is
\[
\hat y_m(x)
=
\hat\mu_{g(x),m}
+
\gamma h_m(\psi_{\mathrm{res}}(x)).
\]

This decomposition deliberately assigns different roles to the two terms.
The hierarchical component captures stable capability patterns that can be
reliably shared under sparse supervision, while the residual predictor only
models instance-specific deviations.

\noindent \textbf{Cost-Aware Routing.}
The acquired query--model pairs provide both quality and serving-cost
observations.
Using the same sparse observation set $\Omega$, we estimate the expected cost
of each observed group--model pair by its sample mean and back off to the
corresponding model-level mean when group-specific observations are
unavailable.
We denote the resulting estimate by $\hat c_{gm}$.

At deployment, \textsc{SaveRouter}  combines the predicted quality and serving cost:
\[
\pi_\lambda(x)
=
\arg\max_{m\in\mathcal{M}}
\left[
\hat y_m(x)
-
\lambda
\frac{\hat c_{g(x),m}}{C_{\max}}
\right],
\]
where $C_{\max}$ is the largest model-level mean cost estimated from the
acquired training pairs, and $\lambda$ controls the quality--cost trade-off.
Both quality and cost estimation use only feedback in $\Omega$.
\section{Experiment}

\subsection{Experimental Setup}
\label{sec:exp_setup}

\noindent \textbf{Benchmarks.}
We evaluate \textsc{SaveRouter} on four routing benchmarks:
LLMRouterBench \citep{li2026llmrouterbench}, Mixinstruct \citep{jiang2023llm},
MMR-Bench \citep{ma2026mmr}, and RouterBench \citep{hu2024routerbench}.
For all benchmarks, we use the same in-domain 20\%/80\% train/test split with
random seed 42.
For query grouping, text benchmarks use frozen
\texttt{all-MiniLM-L6-v2} representations, while MMRBench uses concatenated
CLIP \texttt{ViT-B/16} text and image representations.
The query-level residual predictor uses a separate lightweight feature
representation described in Appendix~\ref{app:implementation}.

\noindent \textbf{Baselines.}
We compare \textsc{SaveRouter} with EmbedLLM \citep{zhuang2025embedllm},
kNN \citep{stripelis2024tensoropera}, OmniRouter \citep{mei2025omnirouter},
RMSoftmax \citep{tsiourvas2025causal}, TRouter \citep{liu2026task},
UniRoute \citep{jitkrittum2026universal},
InferenceDynamics \citep{shi2025inference},
WISERouter \citep{li2026wiserouter}, BaRP \citep{wei2025learning}, and
SemiRouter \citep{wang2026semirouter}.
All methods are evaluated under the same data splits within the \textbf{ORBIT} toolkit
\citep{lai2026orbit}. WISERouter, BaRP, and SemiRouter are paper-derived
reimplementations adapted to this evaluation pipeline.

\noindent \textbf{Evaluation Metrics.}
Our primary metrics account for the model-execution cost of acquiring routing
supervision together with the subsequent model-serving cost, rather than
complete system-level cost.
All supervision-amortized metrics are evaluated within each benchmark and
should not be interpreted as comparable absolute monetary costs across
benchmarks; SA-BEP and SA-CR are invariant to a common positive rescaling of
all cost terms within a benchmark.
Let $Q_b$ and $C_b$ denote the quality and per-query serving cost of the best
single model.
We define the upfront supervision cost $C_0$ as the total model-execution cost
used to acquire routing supervision, and let $c_r$ denote the minimum average
model-serving cost at which the router reaches $Q_b$.
The resulting per-query saving is $\Delta c=C_b-c_r$.
When the target quality is reached and $\Delta c>0$, we define
\[
    \mathrm{SA\mbox{-}BEP}
    =
    \left\lceil\frac{C_0}{\Delta c}\right\rceil,
    \qquad
    \mathrm{SA\mbox{-}CR}(H)
    =
    \frac{C_0+Hc_r}{HC_b}.
    \label{eq:amortized_metrics}
\]
\textbf{SA-BEP} measures the deployment volume required to recover the upfront
supervision cost, while \textbf{SA-CR} measures the cost ratio after amortizing
this cost over $H$ deployment queries.
We report either metric as $\infty$ when the target quality is unreachable or
no positive saving is achieved.

We additionally report \textbf{Peak Score ($P_s$)}, the highest test score
across routing operating points, and \textbf{Cost Ratio (CR)}, the minimum
normalized model-serving cost required to reach $Q_b$.
Higher $P_s$ and lower CR, SA-BEP, and SA-CR are better.

\subsection{\textsc{SaveRouter} Pays Back Earlier without Sacrificing Routing Quality}
\label{sec:main_results}

\begin{table*}[t]
\centering
\caption{
Main results on four routing benchmarks.
Higher $P_s$ is better; lower CR, SA-BEP, and SA-CR@1M are better.
$\infty$ denotes an unreachable target operating point; for
supervision-amortized metrics, it also indicates non-positive per-query saving.
For sparse- or partial-feedback methods, every model invocation used to
acquire feedback is charged separately.
}
\label{tab:main_results}

\footnotesize
\setlength{\tabcolsep}{3.8pt}
\renewcommand{\arraystretch}{1.08}

\resizebox{0.98\textwidth}{!}{
\begin{tabular}{
@{}
l
cccc
cccc
@{}
}
\toprule
&
\multicolumn{4}{c}{\textbf{LLMRouterBench}}
&
\multicolumn{4}{c}{\textbf{Mixinstruct}}
\\[-1pt]
\cmidrule(lr){2-5}
\cmidrule(lr){6-9}

\textbf{Method}
& \makecell{$P_s$ $\uparrow$}
& \makecell{CR $\downarrow$}
& \makecell{SA-BEP $\downarrow$}
& \makecell{SA-CR@1M $\downarrow$}
& \makecell{$P_s$ $\uparrow$}
& \makecell{CR $\downarrow$}
& \makecell{SA-BEP $\downarrow$}
& \makecell{SA-CR@1M $\downarrow$}
\\
\midrule

EmbedLLM
& 0.6094 & 0.5246 & 15.6K & 0.5321
& 0.7488 & 0.9924 & 28.60M & 1.2088
\\

kNN
& 0.6230 & 0.3770 & 11.9K & 0.3845
& 0.7484 & 0.9814 & 11.63M & 1.1977
\\

OmniRouter
& 0.6218 & 0.3564 & 11.5K & 0.3638
& 0.7435 & $\infty$ & $\infty$ & $\infty$
\\

RMSoftmax
& 0.6126 & 0.6328 & 20.2K & 0.6403
& 0.7495 & 0.9893 & 20.19M & 1.2056
\\

TRouter
& 0.6145 & 0.5221 & 15.6K & 0.5295
& 0.7488 & 1.0000 & $\infty$ & $\infty$
\\

UniRoute
& 0.6008 & 0.5583 & 16.8K & 0.5657
& 0.7491 & 1.0000 & $\infty$ & $\infty$
\\

InferenceDyn.
& 0.6172 & 0.4952 & 14.7K & 0.5026
& 0.7491 & 1.0000 & $\infty$ & $\infty$
\\

\addlinespace[2pt]
\midrule
\addlinespace[1pt]

WISERouter
& 0.5513 & $\infty$ & $\infty$ & $\infty$
& 0.7483 & 1.0000 & $\infty$ & $\infty$
\\

BaRP
& 0.6181 & 0.3161 & 114.0K & 0.3941
& 0.7482 & 0.9695 & 44.66M & 2.3329
\\

SemiRouter
& 0.5913 & 0.8509 & 26.5K & 0.8548
& 0.7476 & $\infty$ & $\infty$ & $\infty$
\\

\addlinespace[2pt]
\midrule

\rowcolor{black!4}
\textbf{\textsc{SaveRouter}}
& \textbf{0.6338} & \textbf{0.2320} & \textbf{5.3K} & \textbf{0.2360}
& \textbf{0.7498} & \textbf{0.9359} & \textbf{1.23M} & \textbf{1.0148}
\\

\bottomrule
\end{tabular}
}

\vspace{8pt}

\resizebox{0.98\textwidth}{!}{
\begin{tabular}{
@{}
l
cccc
cccc
@{}
}
\toprule
&
\multicolumn{4}{c}{\textbf{MMR-Bench}}
&
\multicolumn{4}{c}{\textbf{RouterBench}}
\\[-1pt]
\cmidrule(lr){2-5}
\cmidrule(lr){6-9}

\textbf{Method}
& \makecell{$P_s$ $\uparrow$}
& \makecell{CR $\downarrow$}
& \makecell{SA-BEP $\downarrow$}
& \makecell{SA-CR@1M $\downarrow$}
& \makecell{$P_s$ $\uparrow$}
& \makecell{CR $\downarrow$}
& \makecell{SA-BEP $\downarrow$}
& \makecell{SA-CR@1M $\downarrow$}
\\
\midrule

EmbedLLM
& 0.7461 & 1.0000 & $\infty$ & $\infty$
& 0.8053 & 0.9296 & 326.4K & 0.9526
\\

kNN
& 0.7475 & 1.0000 & $\infty$ & $\infty$
& 0.8021 & $\infty$ & $\infty$ & $\infty$
\\

OmniRouter
& 0.7435 & $\infty$ & $\infty$ & $\infty$
& 0.7994 & $\infty$ & $\infty$ & $\infty$
\\

RMSoftmax
& 0.7454 & 1.0350 & $\infty$ & $\infty$
& 0.8050 & 0.9865 & 1.70M & 1.0095
\\

TRouter
& 0.7536 & 0.8396 & 34.1K & 0.8451
& 0.7936 & $\infty$ & $\infty$ & $\infty$
\\

UniRoute
& 0.7430 & $\infty$ & $\infty$ & $\infty$
& 0.8062 & 0.9327 & 341.3K & 0.9557
\\

InferenceDyn.
& 0.7525 & 0.9641 & 152.4K & 0.9696
& 0.8053 & 0.9892 & 2.13M & 1.0122
\\

\addlinespace[2pt]
\midrule
\addlinespace[1pt]

WISERouter
& 0.6796 & $\infty$ & $\infty$ & $\infty$
& 0.7934 & $\infty$ & $\infty$ & $\infty$
\\

BaRP
& 0.7448 & 0.9052 & 640.8K & 0.9660
& 0.8053 & 0.9355 & 3.25M & 1.1449
\\

SemiRouter
& 0.7316 & $\infty$ & $\infty$ & $\infty$
& 0.7636 & $\infty$ & $\infty$ & $\infty$
\\

\addlinespace[2pt]
\midrule

\rowcolor{black!4}
\textbf{\textsc{SaveRouter}}
& \textbf{0.7540} & \textbf{0.7765} & \textbf{18.2K} & \textbf{0.7806}
& \textbf{0.8084} & \textbf{0.6810} & \textbf{42.6K} & \textbf{0.6946}
\\

\bottomrule
\end{tabular}
}
\vspace{-0.5cm}
\end{table*}

\noindent \textbf{Comparison with conventional routers.}
\textsc{SaveRouter} achieves the highest $P_s$ and the lowest CR on all four benchmarks,
despite relying on substantially less supervision than fully supervised
routers.
More importantly, this translates directly into earlier payback.
On LLMRouterBench, \textsc{SaveRouter} reaches break-even after only 5.3K deployment
queries, compared with 11.5K for the fastest conventional baseline.
The gap becomes larger on RouterBench, where \textsc{SaveRouter}
requires 42.6K queries versus 326.4K for EmbedLLM.
Mixinstruct provides an especially informative case: routing quality is already highly saturated across methods, with most
$P_s$ values close to 0.75,
yet \textsc{SaveRouter} reduces SA-BEP from 11.63M for the fastest conventional router
to 1.23M.
These results show that reducing supervision cost can substantially shorten
payback even when serving-time routing quality changes only marginally.

\noindent \textbf{Comparison with sparse-feedback routers.}
\textsc{SaveRouter} also compares favorably with methods explicitly designed for
limited or partial feedback.
WISERouter and SemiRouter frequently fail to reach the target operating point
or to produce positive net savings, resulting in unreachable
supervision-amortized metrics on several benchmarks.
BaRP more consistently reaches competitive routing quality, but its repeated
bandit interactions incur substantially larger supervision expenditure under
our supervision-cost accounting.
Consequently, its SA-BEP is 114.0K, 44.66M, 640.8K, and 3.25M across the four
benchmarks, compared with 5.3K, 1.23M, 18.2K, and 42.6K for \textsc{SaveRouter}.
Thus, sparse feedback alone is not sufficient: how the limited supervision is
acquired and shared across queries is critical to making routing economically
effective.

\subsection{Ablation: What Makes \textsc{SaveRouter} Work?}
\label{sec:ablation}

We ablate both the sparse capability estimator and the acquisition policy on
LLMRouterBench under the same supervision budget of $K=4$.
For the estimator, removing query-level residual correction lowers $P_s$ by
1.63 percentage points and increases CR from 0.2320 to 0.3484, showing that
group-level estimates alone miss important query-specific variation.
Replacing task-aware groups with embedding-based clusters also degrades both
routing quality and cost efficiency, while collapsing all queries into a
single group causes the largest drop in $P_s$, from 0.6338 to 0.5985.

We next isolate the acquisition policy while keeping the downstream estimator
fixed.
Random-$K$, capability-only, and uncertainty-only acquisition all underperform
the proposed UCB policy in either routing quality or serving efficiency.
Notably, Random-$K$ reaches break-even earlier (3.8K vs.\ 5.3K queries), whereas
\textsc{SaveRouter} achieves higher $P_s$ and lower SA-CR@1M, showing that
earliest payback and long-horizon efficiency need not coincide.
Combining capability and uncertainty yields the strongest quality--cost
trade-off, supporting the use of both signals for selecting informative
feedback under a fixed supervision budget.

\begin{figure*}[t]
\centering

\begin{minipage}[t]{0.56\textwidth}
\vspace{0pt}
\centering

\captionof{table}{
Ablation on LLMRouterBench.
}
\label{tab:ablation}

\vspace{2pt}
\resizebox{\linewidth}{!}{
\begin{tabular}{lcccc}
\toprule
Variant
& $P_s$ $\uparrow$
& CR $\downarrow$
& SA-BEP $\downarrow$
& SA-CR@1M $\downarrow$ \\
\midrule

\multicolumn{5}{l}{\textit{(a) Estimator ablations}} \\

w/o Query Residual
& 0.6175
& 0.3484
& 6.2K
& 0.3525 \\

Embedding Groups
& 0.6220
& 0.3229
& 5.5K
& 0.3266 \\

Global Group
& 0.5985
& 0.3190
& 6.8K
& 0.3236 \\

\midrule
\multicolumn{5}{l}{\textit{(b) Acquisition ablations}} \\

Random-$K$
& 0.6183
& 0.2958
& \textbf{3.8K}
& 0.2983 \\

Capability-only
& 0.6291
& 0.2587
& 5.9K
& 0.2630 \\

Uncertainty-only
& 0.6135
& 0.3411
& 4.0K
& 0.3435 \\

\specialrule{0.8pt}{2.5pt}{2pt}

\textbf{\textsc{SaveRouter}}
& \textbf{0.6338}
& \textbf{0.2320}
& 5.3K
& \textbf{0.2360} \\

\bottomrule
\end{tabular}
}
\end{minipage}
\hfill
\begin{minipage}[t]{0.40\textwidth}
\vspace{0pt}
\centering

\includegraphics[width=\linewidth]{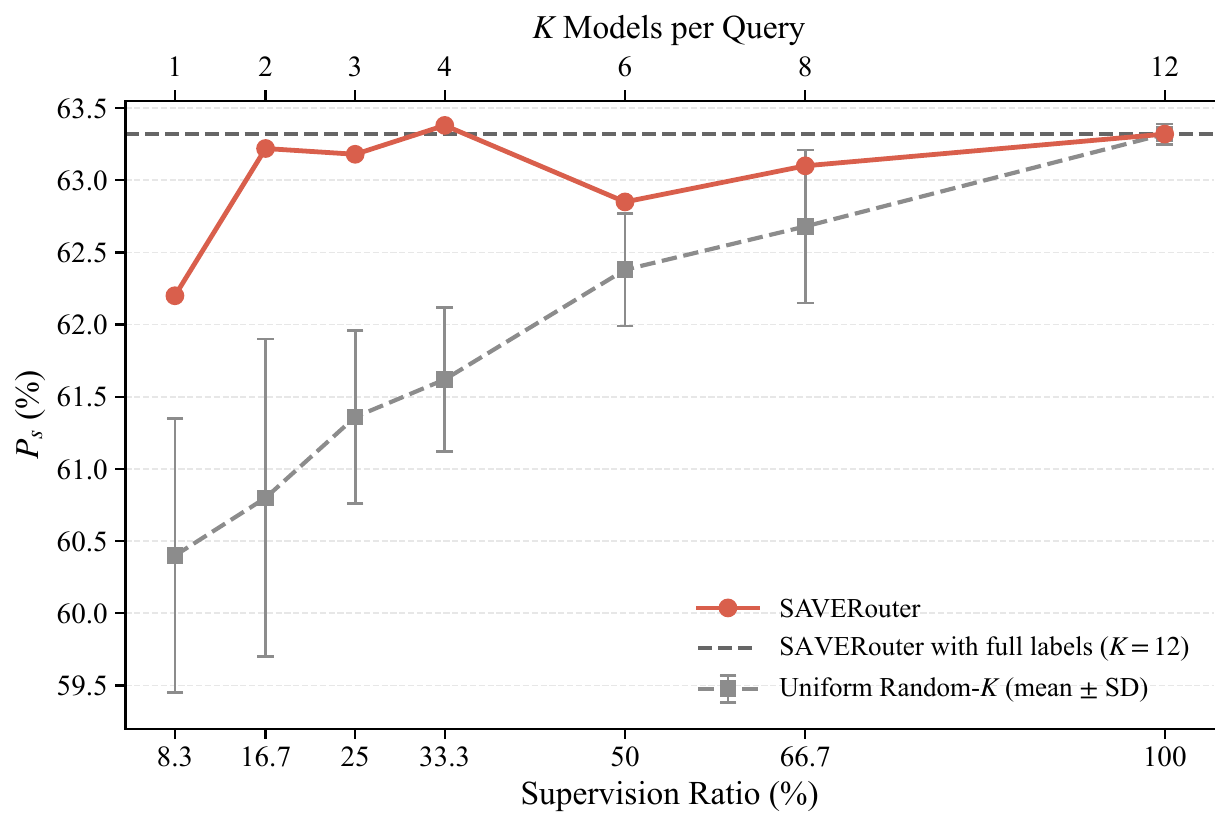}

\vspace{-3pt}
\captionof{figure}{
Effect of supervision budget on routing quality.
}
\label{fig:supervision_quality}

\end{minipage}
\vspace{-0.3cm}
\end{figure*}

\subsection{How Much Supervision Is Enough?}
\label{sec:supervision_analysis}

Figure~\ref{fig:supervision_quality} compares \textsc{SaveRouter} with Uniform
Random-$K$ under the same supervision budget; error bars show the standard
deviation of random selection, and the dashed line denotes full supervision.
With only $K=2$ models evaluated per query (16.7\% supervision), \textsc{SaveRouter}
already approaches its full-supervision accuracy and clearly outperforms
Uniform Random-$K$.
Increasing $K$ further yields only modest gains, indicating rapidly diminishing
returns from additional supervision.

Together with the acquisition ablation in Table~\ref{tab:ablation}, these
results highlight two complementary aspects of supervision efficiency:
effective routing depends not only on \emph{how much} feedback is acquired, but
also on \emph{which} feedback is selected.
Dense supervision is therefore often unnecessary once the acquired observations
are sufficiently informative for downstream routing decisions.

Performance is not strictly monotonic in $K$.
We do not interpret these small fluctuations as evidence that additional
supervision is intrinsically harmful.
Changing the supervision budget changes the observations used to fit the
capability estimator and can consequently alter the learned routing frontier.
The broader trend is therefore more informative than any individual operating
point: routing quality saturates early, while additional supervision provides
limited marginal benefit.
\section{Conclusion}

We introduce \textsc{SaveRouter}, a sparse-supervision routing framework that
learns effective routing policies from limited query--model feedback.
By selectively acquiring informative outcomes and sharing capability
information across related queries, \textsc{SaveRouter} maintains competitive
routing quality with substantially lower supervision expenditure.
Across benchmarks and operating regimes, this results in earlier break-even and
lower supervision-amortized cost.
Our results show that economical LLM routing depends not only on serving-time
efficiency, but also on how much supervision is acquired before deployment.


\bibliography{iclr2027_conference}
\bibliographystyle{iclr2027_conference}

\clearpage
\appendix

\setcounter{figure}{0}
\setcounter{table}{0}
\renewcommand{\thefigure}{A\arabic{figure}}
\renewcommand{\thetable}{A\arabic{table}}

\section*{Appendix}

\section{Cost Accounting with Upfront Supervision}
\label{app:supervision_cost}

Our evaluation accounts for both the upfront cost of acquiring routing
supervision and the model-serving cost incurred after deployment.
This section describes the cost accounting protocol used throughout our
experiments and provides detailed definitions of the supervision-amortized
metrics.

\subsection{Cost Components and Assumptions}
\label{app:cost_accounting}

\noindent \textbf{Supervision acquisition cost.}
We define the upfront supervision cost $C_0$ as the total model-execution cost
of acquiring the feedback used to construct a router.
For each observed query--model pair $(i,m)$, let $a_{im}$ denote the
model-execution cost provided by the corresponding benchmark.
Given an observation set $\Omega$, the total supervision cost is

\[
C_0(\Omega)
=
\sum_{(i,m)\in\Omega} a_{im}.
\label{eq:app_supervision_cost}
\]

This definition charges only model evaluations used to obtain routing
feedback; router training itself is not included in $C_0$.

\noindent \textbf{Fully supervised routers.}
For conventional fully supervised methods, we charge all available training
query--model feedback.
For a complete matrix with $N$ training queries and $M$ candidate models,
the supervision cost is

\[
C_0^{\mathrm{dense}}
=
\sum_{i=1}^{N}\sum_{m=1}^{M} a_{im}.
\]

Thus, all model evaluations required to obtain dense supervision are counted
before deployment.

\noindent \textbf{Sparse-supervision routers.}
For \textsc{SaveRouter}, only the query--model outcomes actually acquired
during sparse supervision are charged.
When $K$ distinct models are evaluated for each training query,
$|\Omega|=NK$, and

\[
C_0^{\mathrm{SAVE}}
=
\sum_{i=1}^{N}
\sum_{m\in\Omega_i}
a_{im},
\qquad
|\Omega_i|=K.
\]

The same accounting principle is applied to sparse- or partial-feedback
baselines, including BaRP, WISERouter, and SemiRouter:
each model invocation used to acquire feedback contributes to $C_0$.

\noindent \textbf{Repeated feedback acquisition.}
Our main experiments use fresh-feedback accounting, where repeated feedback
requests are charged as separate model invocations rather than free cache
lookups.
This corresponds to a setting in which each interaction requests a fresh
model response and therefore incurs a new inference cost
\citep{lai2026dars}.
Accordingly, methods that acquire feedback through repeated interactions are
charged for every such interaction.
We additionally report a cached-feedback sensitivity analysis for BaRP in
Appendix~\ref{app:barp_cache}.

Overall, this protocol measures the supervision expenditure required by each
method while using the same benchmark-provided model costs across methods.

\subsection{Detailed Definitions of SA-BEP and SA-CR}
\label{app:amortized_metrics}

We evaluate supervision-amortized efficiency relative to the best single-model
reference system.
Let $Q_b$ denote the quality of the best single model and $C_b$ its average
per-query serving cost.
For a routing method, let $c_r$ denote the minimum average model-serving cost
among operating points whose routing quality reaches $Q_b$.
The conventional serving-time cost ratio is therefore

\[
\mathrm{CR}
=
\frac{c_r}{C_b}.
\]

The resulting per-query saving is

\[
\Delta c
=
C_b-c_r.
\label{eq:app_delta_c}
\]

\noindent \textbf{Supervision-Amortized Break-Even Point.}
If the router reaches the target quality and $\Delta c>0$, its upfront
supervision cost is recovered after

\[
\mathrm{SA\mbox{-}BEP}
=
\left\lceil
\frac{C_0}{\Delta c}
\right\rceil.
\label{eq:app_sabep}
\]

SA-BEP therefore measures how much deployment traffic is required before the
cumulative serving-time savings compensate for the supervision expenditure
used to construct the router.
A smaller SA-BEP indicates earlier payback.

If the router cannot reach $Q_b$, or if $\Delta c\leq0$, no finite deployment
volume can recover its upfront supervision cost under the target-quality
requirement.
We therefore report SA-BEP as $\infty$ in these cases.

\noindent \textbf{Supervision-Amortized Cost Ratio.}
For a fixed deployment horizon of $H$ queries, the supervision and serving
costs considered in our evaluation sum to

\[
C_{\mathrm{amort}}(H)
=
C_0 + Hc_r,
\]

whereas directly serving the reference model costs $HC_b$.
We define

\[
\mathrm{SA\mbox{-}CR}(H)
=
\frac{C_0+Hc_r}
     {HC_b}.
\label{eq:app_sacr}
\]

A value below one indicates that the serving-time savings have recovered the
upfront supervision expenditure by deployment horizon $H$.
Unless otherwise stated, we report
$\mathrm{SA\mbox{-}CR}@1\mathrm{M}$ using $H=10^6$ deployment queries.

For consistency with SA-BEP, when the target quality is unreachable or
$\Delta c\leq0$, we report SA-CR as $\infty$, indicating that the router does
not reach an economically viable target-quality operating point under this
accounting.

\section{Analytical Properties}
\label{app:theory}

This section provides several simple results that clarify the principles behind
supervision-efficient routing.
We first relate the two supervision-amortized metrics, then characterize
when additional supervision is economically justified. We further formalize
the notion of decision sufficiency.

\subsection{Relationship between SA-BEP and SA-CR}
\label{app:theory_metrics}

Let $C_0$ denote the upfront supervision cost, $C_b$ the per-query cost of the
reference model, and $c_r$ the per-query serving cost of the router at the
target-quality operating point.
The corresponding per-query saving is
\[
\Delta c = C_b-c_r.
\]

\noindent\textbf{Proposition 1.}
Suppose the router reaches the target quality and $\Delta c>0$.
Then
\[
\mathrm{SA\mbox{-}CR}(H)
=
1+
\frac{C_0-H\Delta c}{HC_b}.
\]
Consequently,
\[
\mathrm{SA\mbox{-}CR}(H)\leq 1
\quad\Longleftrightarrow\quad
H\geq \frac{C_0}{\Delta c}.
\]
Therefore,
\[
\mathrm{SA\mbox{-}BEP}
=
\left\lceil
\frac{C_0}{\Delta c}
\right\rceil
\]
is exactly the smallest integer deployment horizon at which the serving-time
savings recover the upfront supervision expenditure.

\noindent\textbf{Proof.}
By definition,
\[
\mathrm{SA\mbox{-}CR}(H)
=
\frac{C_0+Hc_r}{HC_b}.
\]
Since $c_r=C_b-\Delta c$,
\[
\mathrm{SA\mbox{-}CR}(H)
=
\frac{C_0+H(C_b-\Delta c)}{HC_b}
=
1+\frac{C_0-H\Delta c}{HC_b}.
\]
Because $HC_b>0$, the ratio is at most one if and only if
$H\Delta c\geq C_0$, which gives the result.
\hfill$\square$

\noindent\textbf{Interpretation.}
SA-BEP and SA-CR describe the same supervision-amortized economics from two
complementary views.
SA-BEP measures \emph{when} the upfront supervision investment is recovered,
whereas SA-CR measures the resulting cost ratio at a specified deployment
scale.

\subsection{When Is Additional Supervision Worth Its Cost?}
\label{app:theory_supervision}

Consider two routers constructed with different amounts of supervision.
Let
\[
(C_0^{(1)},s_1)
\quad\text{and}\quad
(C_0^{(2)},s_2)
\]
denote their upfront supervision costs and per-query serving costs,
respectively, where both routers satisfy the same target-quality requirement.
Suppose router~2 uses more supervision:
\[
\Delta C_0
=
C_0^{(2)}-C_0^{(1)}
>0.
\]

\noindent\textbf{Proposition 2.}
Additional supervision is economically beneficial over a deployment horizon
$H$ if and only if the serving-cost reduction it induces is large enough to
recover its additional acquisition cost:
\[
H(s_1-s_2)>\Delta C_0.
\]

If
\[
\delta=s_1-s_2>0,
\]
the more heavily supervised router becomes preferable only after
\[
H>
H_{\mathrm{cross}}
=
\frac{\Delta C_0}{\delta}.
\]

If $\delta\leq0$, the additional supervision never reduces the accounted
expenditure.

\noindent\textbf{Proof.}
The accounted expenditures of the two routers after $H$ deployment queries are
\[
L_1(H)=C_0^{(1)}+Hs_1
\]
and
\[
L_2(H)=C_0^{(2)}+Hs_2.
\]
Router~2 is economically preferable when
\[
L_2(H)<L_1(H).
\]
Substituting the definitions gives
\[
C_0^{(2)}-C_0^{(1)}
<
H(s_1-s_2),
\]
which is equivalent to
\[
\Delta C_0<H\delta.
\]
\hfill$\square$

\noindent\textbf{Interpretation.}
This result formalizes the notion of economically over-provisioned supervision.
Additional feedback is worthwhile only when the resulting serving-time
improvement is large enough to repay its acquisition cost within the intended
deployment horizon.
In particular, once routing performance has largely saturated, the marginal
serving benefit $\delta$ can become small, causing $H_{\mathrm{cross}}$ to grow
rapidly even if additional supervision still yields a measurable performance
improvement.

\subsection{Decision Sufficiency for Routing}
\label{app:theory_decision}

Dense supervision aims to estimate the complete query--model performance
matrix.
Routing, however, only requires identifying the model with the highest
cost-adjusted utility.

For a query $x$, define the true utility of model $m$ as
\[
u_m(x)
=
y_m(x)
-
\lambda\frac{c_m(x)}{c_{\max}},
\]
and its estimated utility as
\[
\hat u_m(x)
=
\hat y_m(x)
-
\lambda\frac{c_m(x)}{c_{\max}}.
\]

Let
\[
m^*(x)
=
\arg\max_m u_m(x)
\]
denote the optimal routing decision.

\noindent\textbf{Proposition 3.}
Suppose the quality estimate of each model satisfies
\[
|\hat y_m(x)-y_m(x)|
\leq
\epsilon_m(x).
\]
If for every $m\neq m^*(x)$,
\[
u_{m^*}(x)-u_m(x)
>
\epsilon_{m^*}(x)+\epsilon_m(x),
\]
then the estimated router makes exactly the same decision:
\[
\arg\max_m \hat u_m(x)
=
m^*(x).
\]

In particular, if
\[
|\hat y_m(x)-y_m(x)|\leq\epsilon
\qquad
\forall m,
\]
it is sufficient that the utility margin satisfies
\[
u_{m^*}(x)
-
\max_{m\neq m^*(x)}u_m(x)
>
2\epsilon.
\]

\noindent\textbf{Proof.}
For any $m\neq m^*(x)$,
\[
\hat u_{m^*}(x)-\hat u_m(x)
\geq
u_{m^*}(x)-u_m(x)
-\epsilon_{m^*}(x)-\epsilon_m(x).
\]
Under the stated condition, the right-hand side is strictly positive.
Therefore,
\[
\hat u_{m^*}(x)>\hat u_m(x)
\qquad
\forall m\neq m^*(x),
\]
and the routing decision is unchanged.
\hfill$\square$

\noindent\textbf{Interpretation.}
Accurate reconstruction of every query--model outcome is therefore unnecessary.
Once the estimation error is smaller than the decision margin, further
improving individual capability estimates cannot change the selected model.
This formalizes the distinction between \emph{information completeness} and
\emph{decision sufficiency} used in Section~3.

\section{Implementation Details}
\label{app:implementation}

\subsection{\textsc{SaveRouter} Algorithm}
\label{app:algorithm}

\textsc{SaveRouter} consists of query grouping, sparse feedback acquisition,
structured capability estimation, query-level residual correction, and
cost-aware routing.
The complete procedure is summarized below.

\medskip
\noindent\textbf{Algorithm 1: \textsc{SaveRouter} training and routing.}

\begin{enumerate}

\item \textbf{Fit query groups.}
Using training inputs only, learn a grouping function $\hat g(x)$.
When predictable task labels are available, training queries use their observed
training labels and a classifier is fitted to assign groups to unseen queries.
Otherwise, latent groups are learned from frozen query representations.

\item \textbf{Acquire sparse feedback.}
Initialize the persistent observation mask $O_{im}=0$.
For acquisition pass $p=0,\ldots,K-1$, reset the pass-specific statistics
$\{S_m,n_m,S_{gm},n_{gm}\}$ and shuffle all training queries using seed
$42+1009p$.

For query $x_i$ with group $g_i$, define
\[
\mathcal C_i
=
\{m\in\mathcal M:
m\text{ is available for }x_i,\ O_{im}=0\}.
\]
If $\mathcal C_i$ contains a model with $n_{g_i m}=0$, uniformly sample one
such model.
Otherwise select the model with the largest empirical-Bayes UCB score,
\[
\bar\mu_m
=
\frac{S_m+\alpha_0}
     {n_m+\alpha_0+\beta_0},
\qquad
\tilde\mu_{gm}
=
\frac{S_{gm}+\tau_0\bar\mu_m}
     {n_{gm}+\tau_0},
\]
\[
m_i
=
\arg\max_{m\in\mathcal C_i}
\left[
\tilde\mu_{g_i m}
+
\beta_{\mathrm{ucb}}
\sqrt{
\frac{\log(\sum_j n_{g_i j}+1)}
     {\max(n_{g_i m},1)}
}
\right].
\]
Only after $m_i$ is selected do we reveal $y_{im_i}$, update the current-pass
statistics, and set $O_{im_i}=1$.
After $K$ passes, up to $K$ distinct model outcomes are acquired for each
training query, subject to model availability.

\item \textbf{Estimate group--model capability.}
Let
\[
\Omega=\{(i,m):O_{im}=1\}
\]
denote all acquired query--model pairs.
Using all observations in $\Omega$, fit the two-way additive Ridge model
\[
y_{im}
=
b+u_{g_i}+v_m+\epsilon_{im}.
\]
Its structured prior is
\[
\pi_{gm}
=
\operatorname{clip}
\left(
\hat b+\hat u_g+\hat v_m,\,
0,1
\right).
\]
Aggregating statistics over all acquisition passes, the final group--model
estimate is
\[
\hat\mu_{gm}
=
\frac{
S^{\Omega}_{gm}
+
\tau\pi_{gm}
}{
n^{\Omega}_{gm}+\tau
}.
\]

\item \textbf{Fit query-level residuals.}
For each acquired pair $(i,m)$, construct the leave-one-out group baseline
\[
\hat\mu^{-i}_{g_i m}
=
\frac{
S^{\Omega}_{g_i m}-y_{im}
+
\tau\pi_{g_i m}
}{
n^{\Omega}_{g_i m}-1+\tau
},
\]
and residual target
\[
e_{im}
=
y_{im}-\hat\mu^{-i}_{g_i m}.
\]
For every model with at least eight acquired observations, fit an independent
linear Ridge predictor $h_m$ from contextual features
$\psi_{\mathrm{res}}(x)$ to $e_{im}$.
The final quality estimate is
\[
\hat y_m(x)
=
\hat\mu_{\hat g(x),m}
+
\gamma h_m(\psi_{\mathrm{res}}(x)).
\]
Models with fewer than eight acquired observations use only the group-level
estimate.
The final $\hat y_m(x)$ is not clipped to $[0,1]$.

\item \textbf{Estimate serving cost from sparse observations.}
Serving-cost estimation uses only the same acquired pairs $\Omega$.
Let $c_{im}$ denote the observed serving cost for pair $(i,m)$.
For a group--model pair with sampled observations, we use
\[
\hat c_{gm}
=
\frac{
\sum_{i:(i,m)\in\Omega,\ g_i=g} c_{im}
}{
n^{\Omega}_{gm}
}.
\]
When a group--model pair has no sampled cost observation, we back off to the
model-level mean computed from sampled pairs,
\[
\bar c_m
=
\frac{
\sum_{i:(i,m)\in\Omega} c_{im}
}{
\sum_i \mathbbm{1}[(i,m)\in\Omega]
}.
\]
Thus, unobserved quality and cost entries outside $\Omega$ are not used to
construct the router.

\item \textbf{Route new queries.}
For a new query $x$, predict its group $\hat g(x)$ and compute all model-quality
and cost estimates.
For a cost weight $\lambda$, route to a single model using
\[
\pi_\lambda(x)
=
\arg\max_{m\in\mathcal M}
\left[
\hat y_m(x)
-
\lambda
\frac{\hat c_{\hat g(x),m}}
     {C_{\max}}
\right],
\]
where
\[
C_{\max}
=
\max_m \bar c_m.
\]

\end{enumerate}

\medskip
\noindent\textbf{Feedback-revelation protocol.}
During acquisition, \textsc{SaveRouter} knows whether a candidate model is
available for a query, but does not observe its quality or serving cost until
that model is selected.
The persistent mask prevents the same query--model pair from being acquired in
multiple passes.
The acquisition policy uses only quality feedback revealed earlier in the
current pass; the final additive prior, residual predictors, and serving costs
are not used to choose acquisition actions.

\subsection{Hyperparameters and Query Grouping}
\label{app:hyperparameters}

\noindent\textbf{Core hyperparameters.}
Table~\ref{tab:saverouter_hparams} lists the hyperparameters used in the main
experiments.
Most values are shared across benchmarks.
The supervision budget $K$ is specified by each experiment and is therefore
not treated as a globally shared hyperparameter.

\begin{table}[t]
\centering
\caption{Core hyperparameters of \textsc{SaveRouter}.}
\label{tab:saverouter_hparams}
\begin{tabular}{lcc}
\toprule
Hyperparameter & Symbol & Value \\
\midrule
Global prior parameter
& $\alpha_0$ & 1 \\
Global prior parameter
& $\beta_0$ & 1 \\
Acquisition prior strength
& $\tau_0$ & 10 \\
UCB exploration coefficient
& $\beta_{\mathrm{ucb}}$ & 0.35 \\
Additive-prior Ridge weight
& $\lambda_{\mathrm{prior}}$ & 10 \\
Final shrinkage strength
& $\tau$ & 40 \\
Residual Ridge weight
& $\lambda_{\mathrm{ctx}}$ & 100 (200 on LLMRouterBench) \\
Residual scaling
& $\gamma$ & 2 \\
Minimum residual observations
& -- & 8 \\
\bottomrule
\end{tabular}
\end{table}

The additive prior uses Ridge regression with an intercept and no explicit
zero-sum constraint on group or model effects.
It is fitted once after all $K$ acquisition passes using the complete sparse
observation set $\Omega$.

\noindent\textbf{Query grouping.}
Grouping is learned exclusively from the training split.
We distinguish the representation used for grouping,
$\phi_{\mathrm{grp}}(x)$, from the contextual representation used by the
residual predictor, $\psi_{\mathrm{res}}(x)$.

The default grouping strategy first checks whether task labels are available
on the training data.
If there are at most $128$ classes and a classifier fitted from frozen input
representations reaches at least $0.8$ training accuracy, the training labels
are used as groups.
A logistic classifier with
$C=10$, \texttt{class\_weight=balanced}, and random seed $42$ is then used to
assign groups to unseen queries.
Training examples use their observed training labels, whereas test examples
use only classifier predictions.

Otherwise, SAVERouter uses latent grouping with MiniBatchKMeans.
We use random seed $42$, $n_{\mathrm{init}}=3$, and batch size
$\min(2048,N)$.
Unless explicitly specified, the number of groups is
\[
G
=
\operatorname{clip}
\left(
\operatorname{round}\left(\frac{\sqrt N}{2}\right),
4,32
\right).
\]
New queries are assigned using the learned nearest-centroid predictor.

\begin{table}[t]
\centering
\caption{Query grouping used in the main experiments.}
\label{tab:saverouter_grouping}
\begin{tabular}{lll}
\toprule
Benchmark & Training groups & New-query assignment \\
\midrule
LLMRouterBench
& 10 task groups
& Logistic classifier \\

Mixinstruct
& Single global group
& Single group \\

MMRBench
& 7 dataset groups
& Logistic classifier \\

RouterBench
& 85 task groups
& Logistic classifier \\
\bottomrule
\end{tabular}
\end{table}

\noindent\textbf{Grouping representations.}
For text benchmarks, $\phi_{\mathrm{grp}}(x)$ is the 384-dimensional
\texttt{all-MiniLM-L6-v2} representation.
For MMRBench, we concatenate the CLIP \texttt{ViT-B/16} text and image
representations into a 1024-dimensional vector.
All grouping representations are $\ell_2$-normalized before classification or
clustering.

\noindent\textbf{Residual representations.}
The residual predictor uses a separate representation from query grouping.
For text benchmarks, $\psi_{\mathrm{res}}(x)$ concatenates word TF--IDF
features with $1$--$2$ grams and character-level \texttt{char\_wb} TF--IDF
features with $3$--$5$ grams.
The corresponding minimum document frequencies are $2$ and $3$, and each
vocabulary contains at most $30{,}000$ features.
We use sublinear term frequency.

For MMRBench, the normalized 1024-dimensional CLIP representation is
additionally concatenated with the TF--IDF features.
All vocabularies and residual predictors are fitted only on the training split.

\noindent\textbf{Acquisition randomness.}
Each acquisition pass independently shuffles all training queries with seed
\[
42+1009p,
\]
where $p$ is the pass index.
Previously unseen group--model pairs are uniformly sampled before applying the
UCB rule.
Score ties are broken using an infinitesimal random perturbation.
No additional top-up observations are acquired beyond the specified budget
$K$.

\noindent\textbf{Cost-aware operating points.}
The standard routing policy is evaluated over
\[
\lambda
\in
\{0\}
\cup
\operatorname{logspace}(10^{-4},10^3,200),
\]
yielding $201$ cost weights.
For offline construction of the quality--cost frontier, we additionally
evaluate cost-threshold policies and incremental predicted-quality-gain versus
incremental predicted-cost gating policies derived from the same trained
quality and cost estimators.

Test outcomes are used only to evaluate these fixed candidate policies and
construct the benchmark frontier; they are never used to train query groups,
select sparse feedback, fit capability estimates, estimate costs, or train
residual predictors.
The resulting frontier is used to report
\textbf{Peak Score ($P_s$)}, CR, SA-BEP, and SA-CR.

\section{Detailed Experimental Setup}
\label{app:experimental_setup}

\subsection{Benchmark Statistics}
\label{app:benchmark_details}

All experiments use an in-domain 20\%/80\% train/test split with random
seed 42.
Table~\ref{tab:benchmark_stats} summarizes the resulting benchmark statistics.
The number of groups corresponds to the grouping used by \textsc{SaveRouter}
in the main experiments; for Mixinstruct, we explicitly use a single global
group.

\begin{table}[t]
\centering
\caption{
Statistics of the routing benchmarks used in our experiments.
Available train pairs count query--model pairs for which both quality and cost
fields are available.
}
\label{tab:benchmark_stats}
\small
\setlength{\tabcolsep}{4.2pt}
\begin{tabular}{lrrrrrr}
\toprule
Benchmark
& \#Queries
& \#Train
& \#Test
& \#Models
& \#Groups
& Train Pairs \\
\midrule
LLMRouterBench
& 12,446 & 2,489 & 9,957 & 12 & 10
& 29,868 / 29,868 \\

Mixinstruct
& 110,000 & 22,000 & 88,000 & 12 & 1
& 264,000 / 264,000 \\

MMRBench
& 10,370 & 2,074 & 8,296 & 10 & 7
& 20,149 / 20,740 \\

RouterBench
& 36,497 & 7,299 & 29,198 & 11 & 85
& 80,289 / 80,289 \\
\bottomrule
\end{tabular}
\end{table}

LLMRouterBench, Mixinstruct, and RouterBench contain complete training
query--model matrices after preprocessing.
MMRBench contains 591 unavailable training pairs and 2,449 unavailable test
pairs; all methods respect these availability masks during training and
evaluation.

\subsection{Baseline Configurations}
\label{app:baseline_config}

All methods are evaluated within the same ORBIT pipeline and use identical
train/test splits.
EmbedLLM, kNN, OmniRouter, RMSoftmax, TRouter, UniRoute, and
InferenceDynamics use their corresponding ORBIT implementations and
configurations.
WISERouter, BaRP, and SemiRouter are paper-derived reimplementations adapted
to the same evaluation protocol rather than executions of the authors'
original code.

Table~\ref{tab:baseline_config} summarizes the key configurations used in our
experiments.
Dense baselines use all available quality feedback in the training split.
For MMRBench, this means all available query--model pairs rather than a
complete $N_{\mathrm{train}}M$ matrix.

\begin{table}[t]
\centering
\caption{Key configurations of the compared routing methods.}
\label{tab:baseline_config}
\small
\setlength{\tabcolsep}{5pt}
\begin{tabular}{lp{0.72\linewidth}}
\toprule
Method & Configuration \\
\midrule

EmbedLLM
& $\alpha=0.001$; 30 epochs; learning rate $10^{-3}$. \\

kNN
& $k=300$. \\

OmniRouter
& top-$k=50$; $\gamma=\delta=0.6$; 50 epochs. \\

RMSoftmax
& 30 linearly spaced cost weights from $0$ to $1000$; 300 epochs. \\

TRouter
& hidden dimension 256; dropout 0.1; temperature 0.07; 20 epochs. \\

UniRoute
& 10 clusters; mapping network trained for 5 epochs. \\

InferenceDynamics
& rank decay 0.8; cost penalty 0; profile source \texttt{eval\_name}. \\

WISERouter
& WR-Online with 16 contexts and one exploration pass, corresponding to one
interaction per training query. \\

BaRP
& 100 training epochs, corresponding to 100 selected-arm feedback interactions
per training query in our implementation; batch size 32; 101 inference
preference points. \\

SemiRouter
& three anchors with all available labels and up to two non-anchor labels per
query; EMB variant; 20 backbone epochs and 2 adapter epochs. \\

\bottomrule
\end{tabular}
\end{table}

Operating points follow the evaluation protocol of each method rather than a
shared \textsc{SaveRouter}-specific grid.
EmbedLLM, kNN, OmniRouter, TRouter, UniRoute, InferenceDynamics, and
SemiRouter use ORBIT's predicted-cost threshold frontier.
RMSoftmax evaluates 30 cost weights, WISERouter evaluates 64 workload
budgets, and BaRP evaluates 101 quality--cost preference settings.

Our ORBIT adaptation of InferenceDynamics uses the recorded
\texttt{eval\_name} to construct task profiles, whereas \textsc{SaveRouter}
predicts the group of an unseen query from its input.
We note this difference in task-side information when interpreting the
comparison.

\subsection{Cost Configuration}
\label{app:cost_config}

The cost fields used by ORBIT are benchmark-specific and do not share a common
absolute monetary scale.
We therefore compare routing costs only within each benchmark.

\noindent\textbf{Benchmark cost fields.}
For LLMRouterBench, the original cost records are preprocessed to repair a
subset of zero-valued entries using token-price or task-level median
information, and are subsequently normalized by the maximum dataset cost.
Mixinstruct uses model-level cost proxy values specified by the benchmark
loader and normalizes them by the maximum proxy value of 16.
MMRBench uses the numerical model-cost fields provided in its data files,
while RouterBench uses the cost fields contained in the original records.

\noindent\textbf{Supervision acquisition cost.}
Supervision cost is accumulated from the benchmark cost field associated with
each model invocation used to obtain quality feedback.
Dense baselines are charged for all available training query--model pairs.
For \textsc{SaveRouter}, all main-table experiments use $K=4$, so that up to
four distinct model outcomes are acquired for each training query, subject to
benchmark availability.

WISERouter and SemiRouter are charged according to the feedback acquired by
their respective protocols.
For BaRP, our main evaluation uses fresh-feedback accounting: repeated
selected-arm interactions are counted as separate model calls rather than free
cache accesses.
We additionally report cached-feedback results in
Appendix~\ref{app:barp_cache}.

\noindent \textbf{Quality and cost information.}
Each acquired query--model pair $(i,m)\in\Omega$ reveals both its quality
feedback $y_{im}$ and realized serving cost $c_{im}$.
\textsc{SaveRouter} uses only these acquired pairs for both capability and
serving-cost estimation; quality and cost entries outside $\Omega$ are
unavailable during router training.
Thus, the supervision budget jointly determines the quality and cost
observations available to the router.

\noindent\textbf{Serving-time cost.}
For each routing operating point, serving cost is computed from the benchmark
cost of the model actually selected for each test query, rather than from the
router's predicted cost.
The reference quality $Q_b$ and cost $C_b$ are obtained from the best single
model on the same test split.
We report $\mathrm{SA\mbox{-}CR}@1\mathrm{M}$ with $H=10^6$.
Because cost definitions and normalizations differ across benchmarks,
absolute costs should not be compared or aggregated across datasets.

\section{Additional Experiments}
\label{app:additional_experiments}

\subsection{Effect of Supervision Budget across Benchmarks}
\label{app:k_scaling}

We further study how \textsc{SaveRouter} behaves under different supervision
budgets.
For each benchmark, we vary the number of acquired model outcomes per query
while keeping all non-budget hyperparameters fixed.
The supervision ratio is computed relative to the number of available
training query--model pairs; therefore, on MMRBench it is not simply $K/M$
because the underlying matrix contains missing entries.

Table~\ref{tab:k_scaling} reports Peak Score ($P_s$), the target-quality
Cost Ratio (CR), supervision acquisition cost $C_0$, SA-BEP, and
SA-CR@1M.
All results use the same seed-42 train/test split as the main experiments.

\begin{table}[t]
\centering
\caption{
Effect of supervision budget $K$ across four routing benchmarks.
Higher $P_s$ is better; lower CR, $C_0$, SA-BEP, and SA-CR@1M are better.
``--'' indicates that the router does not reach the best-single-model quality
target, so the corresponding supervision-amortized metrics are undefined.
}
\label{tab:k_scaling}
\small
\setlength{\tabcolsep}{4.3pt}
\begin{tabular}{lrrrrrrr}

\toprule
Benchmark
& $K$
& Sup. (\%)
& $P_s$ $\uparrow$
& CR $\downarrow$
& $C_0$ $\downarrow$
& SA-BEP $\downarrow$
& SA-CR@1M $\downarrow$ \\
\midrule

LLMRouterBench
& 1  & 8.33   & 0.621874 & 0.2395 & 44.64  & 1,311 & 0.2405 \\
& 2  & 16.67  & 0.631415 & 0.2357 & 85.21  & 2,489 & 0.2376 \\
& 3  & 25.00  & 0.630160 & 0.2228 & 146.69 & 4,213 & 0.2261 \\
& 4  & 33.33  & 0.633825   & 0.2320 & 181.09 & 5.3K  & 0.2360 \\
& 6  & 50.00  & 0.627699 & 0.2180 & 241.15 & 6,884 & 0.2234 \\
& 8  & 66.67  & 0.627147 & 0.2260 & 279.86 & 8,071 & 0.2322 \\
& 12 & 100.00 & 0.630913 & 0.2232 & 333.01 & 9,569 & 0.2306 \\

\midrule

Mixinstruct
& 1  & 8.33   & 0.749016 & 0.9513 & 14,850.06  & 406,582   & 0.9711 \\
& 2  & 16.67  & 0.749304 & 0.9447 & 29,656.44  & 715,324   & 0.9843 \\
& 3  & 25.00  & 0.749366 & 0.9426 & 44,312.56  & 1,029,726 & 1.0017 \\
& 4  & 33.33  & 0.749834   & 0.9359 & 58,642.25  & 1.23M     & 1.0148 \\
& 6  & 50.00  & 0.749741 & 0.9375 & 86,564.94  & 1,845,546 & 1.0529 \\
& 8  & 66.67  & 0.749649 & 0.9416 & 114,141.69 & 2,605,865 & 1.0938 \\
& 12 & 100.00 & 0.749593 & 0.9453 & 162,250.00 & 3,953,001 & 1.1616 \\

\midrule

MMRBench
& 1 & 10.29 & 0.752290 & 0.9637 & 11.95 & 40,156 & 0.9651 \\
& 2 & 20.59 & 0.754098 & 0.8144 & 20.73 & 13,632 & 0.8169 \\
& 3 & 30.88 & 0.745902 & 0.8456 & 27.50 & 21,733 & 0.8489 \\
& 4 & 41.17 & 0.754016   & 0.7765 & 33.48 & 18.2K  & 0.7806 \\
& 5 & 51.47 & 0.747589 & 0.7837 & 38.74 & 21,861 & 0.7884 \\
& 6 & 61.76 & 0.748433 & 0.7638 & 41.60 & 21,498 & 0.7689 \\
& 7 & 72.05 & 0.747348 & 0.7367 & 42.77 & 19,829 & 0.7419 \\
& 8 & 82.35 & 0.748433 & 0.7448 & 43.96 & 21,031 & 0.7502 \\
& 9 & 92.64 & 0.748433 & 0.7514 & 44.60 & 21,895 & 0.7568 \\

\midrule

RouterBench
& 1  & 9.09   & 0.803862 & --     & 37.62  & --     & --     \\
& 2  & 18.18  & 0.806506 & 0.6903 & 61.63  & 25,041 & 0.6981 \\
& 3  & 27.27  & 0.806035 & 0.7396 & 87.17  & 42,126 & 0.7506 \\
& 4  & 36.36  & 0.808397   & 0.6810 & 108.23 & 42.6K  & 0.6946 \\
& 6  & 54.55  & 0.809030 & 0.6509 & 131.51 & 47,404 & 0.6674 \\
& 8  & 72.73  & 0.809269 & 0.6417 & 150.08 & 52,707 & 0.6606 \\
& 11 & 100.00 & 0.807634 & 0.6549 & 182.58 & 66,575 & 0.6779 \\

\bottomrule
\end{tabular}
\end{table}

The results reveal a consistent trade-off between supervision expenditure and
routing performance.
Increasing $K$ provides more observations, but the resulting gains are not
monotonic.
For example, LLMRouterBench already reaches near-peak routing quality at
$K=2$--$4$, while additional supervision mainly increases acquisition cost
and delays break-even.
Mixinstruct exhibits an even stronger saturation effect: $P_s$ varies only
marginally across budgets, whereas SA-BEP increases substantially as more
supervision is acquired.

MMRBench and RouterBench further show that the budget minimizing serving-time
CR need not yield the earliest payback.
On MMRBench, $K=7$ gives the lowest CR among the evaluated settings, whereas
$K=2$ gives the earliest SA-BEP.
On RouterBench, the lowest CR is attained at $K=8$, while the earliest finite
SA-BEP occurs at $K=2$.
Thus, additional supervision can improve serving-time efficiency without
necessarily producing earlier payback.

Overall, these results show that routing quality and serving-time cost often
saturate well before full supervision is acquired.
The appropriate supervision budget therefore depends on both the serving-time
benefit obtained from additional feedback and its upfront acquisition cost,
rather than on routing quality alone.

MMRBench is evaluated up to $K=9$, corresponding to 92.64\% of the available
training query--model pairs, so its final row should not be interpreted as
full supervision.

\subsection{Additional Evaluation on RouterEval}
\label{app:routereval}

We further evaluate \textsc{SaveRouter} on RouterEval as an additional
benchmark beyond the four datasets used in the main experiments.
We compare against 17 routing baselines, including MIRT
\citep{song2025irt}, AvengersPro \citep{zhang2025beyond},
and EquiRouter \citep{lai2026routing}, and follow the
same evaluation protocol as in the main experiments.
Table~\ref{tab:routereval} reports $P_s$, CR, SA-BEP, and SA-CR@1M.

\begin{table*}[t]
\centering
\caption{
Results on RouterEval.
\textsc{SaveRouter} uses $K=108$.
Higher $P_s$ is better; lower CR, SA-BEP, and SA-CR@1M are better.
$\infty$ denotes an unreachable target operating point.
}
\label{tab:routereval}
\small
\setlength{\tabcolsep}{7.0pt}
\renewcommand{\arraystretch}{1.03}
\begin{tabular}{lcccc}
\toprule
Method
& $P_s$ $\uparrow$
& CR $\downarrow$
& SA-BEP $\downarrow$
& SA-CR@1M $\downarrow$ \\
\midrule

\multicolumn{5}{l}{\textit{Dense-supervision routers}} \\

EmbedLLM
& 0.776135
& $\infty$
& $\infty$
& $\infty$ \\

kNN
& 0.795418
& 0.816906
& 4,772,514
& 1.6907 \\

MLP
& 0.776705
& $\infty$
& $\infty$
& $\infty$ \\

SVM
& 0.718650
& $\infty$
& $\infty$
& $\infty$ \\

RouteLLM-MF
& 0.780365
& $\infty$
& $\infty$
& $\infty$ \\

GraphRouter
& 0.726733
& $\infty$
& $\infty$
& $\infty$ \\

MIRT
& 0.732729
& $\infty$
& $\infty$
& $\infty$ \\

OmniRouter
& 0.773745
& $\infty$
& $\infty$
& $\infty$ \\

AvengersPro
& 0.772553
& $\infty$
& $\infty$
& $\infty$ \\

RMSoftmax
& 0.791400
& 0.935771
& 13,604,778
& 1.8096 \\

EquiRouter
& 0.785997
& 0.822741
& 4,929,617
& 1.6966 \\

TRouter
& 0.798898
& 0.791405
& 4,189,065
& 1.6652 \\

UniRoute
& 0.763016
& $\infty$
& $\infty$
& $\infty$ \\

InferenceDynamics
& 0.804803
& 0.799087
& 4,349,239
& 1.6729 \\

\midrule
\multicolumn{5}{l}{\textit{Sparse- or partial-feedback routers}} \\

WISERouter
& 0.520140
& $\infty$
& $\infty$
& $\infty$ \\

BaRP
& 0.683279
& $\infty$
& $\infty$
& $\infty$ \\

SemiRouter
& 0.683451
& $\infty$
& $\infty$
& $\infty$ \\

\specialrule{0.8pt}{2.5pt}{2pt}

\textbf{\textsc{SaveRouter} ($K=108$)}
& \textbf{0.805597}
& \textbf{0.778155}
& \textbf{1,876,023}
& \textbf{1.1943} \\

\bottomrule
\end{tabular}
\end{table*}

\textsc{SaveRouter} achieves the highest peak routing quality on RouterEval,
with $P_s=0.805597$, slightly above InferenceDynamics at 0.804803.
The difference is only 0.079 percentage points, indicating that the main gain
on this benchmark comes not from substantially higher peak quality, but from
preserving strong routing quality while improving cost efficiency.

Among dense baselines that reach the best-single-model quality $Q_b$,
TRouter attains the lowest CR at 0.791405 and the earliest break-even at
4.19M deployment queries.
\textsc{SaveRouter} reduces CR further to 0.778155 and reaches break-even
after 1.88M queries, reducing SA-BEP by approximately 55.2\%.
Its SA-CR@1M is also reduced from 1.6652 to 1.1943.
Although this value remains above one at one million queries, the lower
SA-BEP shows that the upfront supervision expenditure is recovered
substantially earlier.

The sparse- or partial-feedback baselines do not reach $Q_b$ on RouterEval,
leading to unreachable target-quality operating points.
This highlights that reducing supervision alone is insufficient if the
resulting feedback does not preserve enough information for high-quality
routing.

Overall, RouterEval exhibits the same pattern as the main benchmarks:
\textsc{SaveRouter} preserves competitive routing quality while improving
both serving-time efficiency and supervision-amortized payback.

\subsection{Sensitivity to Training-Set Size}
\label{app:train_ratio}

We further examine how \textsc{SaveRouter} behaves as the amount of training
data increases on LLMRouterBench.
We vary the train/test split from 20\%/80\% to 80\%/20\% and compare against
five dense baselines under the same split.

\begin{table}[t]
\centering
\caption{
Sensitivity to the training-set fraction on LLMRouterBench.
``Best dense'' denotes the strongest dense baseline for the corresponding
metric among EmbedLLM, kNN, OmniRouter, TRouter, and InferenceDynamics.
$\Delta P_s$ is the absolute improvement of \textsc{SaveRouter} in percentage
points, and $\Delta$CR is the relative reduction in CR.
}
\label{tab:train_ratio}
\small
\setlength{\tabcolsep}{4.2pt}
\renewcommand{\arraystretch}{1.05}
\begin{tabular}{lcccccc}
\toprule
Train/Test
& \multicolumn{3}{c}{$P_s$ $\uparrow$}
& \multicolumn{3}{c}{CR $\downarrow$} \\
\cmidrule(lr){2-4}
\cmidrule(lr){5-7}
& \textsc{SaveRouter}
& Best Dense
& $\Delta P_s$ (pp)
& \textsc{SaveRouter}
& Best Dense
& $\Delta$CR \\
\midrule

20/80
& 0.6338
& 0.6230
& +1.08
& 0.2320
& 0.3564
& 34.9\% \\

40/60
& 0.634842
& 0.624665
& +1.018
& 0.1888
& 0.3275
& 42.4\% \\

60/40
& 0.633461
& 0.626029
& +0.743
& 0.2137
& 0.2965
& 27.9\% \\

80/20
& 0.641566
& 0.636145
& +0.542
& 0.1412
& 0.2275
& 38.0\% \\

\bottomrule
\end{tabular}
\end{table}

\textsc{SaveRouter} maintains a higher peak score and a lower target-quality
serving cost than the strongest dense baseline across all four train/test
splits.
The improvement in $P_s$ ranges from 0.542 to 1.08 percentage points, while
the reduction in CR ranges from 27.9\% to 42.4\%.
Thus, the serving-time advantage is not specific to the default 20\% training
split.

More importantly, increasing the amount of training data exhibits strong
diminishing returns.
With only 20\% of the data, \textsc{SaveRouter} already reaches
$P_s=0.6338$.
Increasing the training fraction to 80\% raises $P_s$ to 0.6416, an absolute
gain of only 0.774 percentage points.
Over the same change, the supervision acquisition cost increases from 181.09
to 834.13, while SA-BEP increases from approximately 5.3K to 22.7K deployment
queries.
Hence, substantially more supervision produces only a modest improvement in
peak routing quality while requiring much longer to recover its upfront cost.

The trend is also not monotonic.
The 40\% training split achieves a lower CR than the 60\% split despite using
less training data, showing that additional supervision does not necessarily
translate into a better serving-time operating point.
Taken together, these results reinforce the central observation that more
supervision can yield only marginal routing improvements while substantially
increasing the expenditure that must be amortized after deployment.

\subsection{Hyperparameter Robustness at K=4}
\label{app:hyperparameter_robustness}

We further examine the sensitivity of \textsc{SaveRouter} to its main
regularization parameters on LLMRouterBench under the same $K=4$ supervision
budget.
All configurations use the same 33.33\% supervision ratio and the same
acquisition cost $C_0=181.09$.
The main configuration uses
$(\gamma,\tau,\lambda_{\mathrm{ctx}})=(2,40,200)$.

\begin{table}[t]
\centering
\caption{
Hyperparameter robustness on LLMRouterBench with fixed $K=4$ supervision.
Higher $P_s$ is better; lower CR, SA-BEP, and SA-CR@1M are better.
}
\label{tab:hyperparameter_robustness}
\small
\setlength{\tabcolsep}{5.5pt}
\begin{tabular}{lrrrr}

\toprule
Configuration
& $P_s$ $\uparrow$
& CR $\downarrow$
& SA-BEP $\downarrow$
& SA-CR@1M $\downarrow$ \\
\midrule

Main:
$\gamma=2,\tau=40,\lambda_{\mathrm{ctx}}=200$
& \textbf{0.633825} & 0.2320 & 5,263 & 0.2360 \\

$\lambda_{\mathrm{ctx}}=100$
& 0.631365 & 0.2182 & 5,171 & 0.2223 \\

$\lambda_{\mathrm{ctx}}=50$
& 0.625791 & 0.2208 & 5,188 & 0.2248 \\

$\gamma=1$
& 0.633373 & 0.2517 & 5,402 & 0.2558 \\

$\gamma=4$
& 0.624385 & 0.2430 & 5,340 & 0.2471 \\

$\tau=20$
& 0.628955 & \textbf{0.2024} & \textbf{5,068} & \textbf{0.2065} \\

$\tau=80$
& 0.632068 & 0.2240 & 5,209 & 0.2280 \\

\bottomrule
\end{tabular}
\end{table}

The results show that \textsc{SaveRouter} remains effective across the tested
regularization settings.
Peak quality varies within approximately one percentage point, while all
configurations that reach the target retain finite SA-BEP and SA-CR.

Different metrics nevertheless favor different parameter settings.
The main configuration with $\lambda_{\mathrm{ctx}}=200$ achieves the highest
$P_s$, whereas $\tau=20$ yields the lowest CR, earliest SA-BEP, and lowest
SA-CR@1M.
This again illustrates that the configuration maximizing routing quality need
not minimize serving cost or supervision-amortized expenditure.

Reducing $\lambda_{\mathrm{ctx}}$ from 200 to 100 slightly lowers peak quality
but improves CR and payback, while stronger or weaker residual scaling changes
the operating trade-off without causing a large collapse in routing quality.
Overall, the qualitative behavior of \textsc{SaveRouter} is not tied to a
single regularization setting, although the preferred configuration depends
on whether peak quality, serving efficiency, or earlier payback is prioritized.

The $\gamma$ and $\tau$ sensitivity runs were conducted with
$\lambda_{\mathrm{ctx}}=100$; they are therefore reported as robustness checks
rather than one-at-a-time perturbations of the main configuration.

\begin{figure*}[t]
    \centering

    \begin{subfigure}[t]{0.49\textwidth}
        \centering
        \includegraphics[width=\linewidth]
        {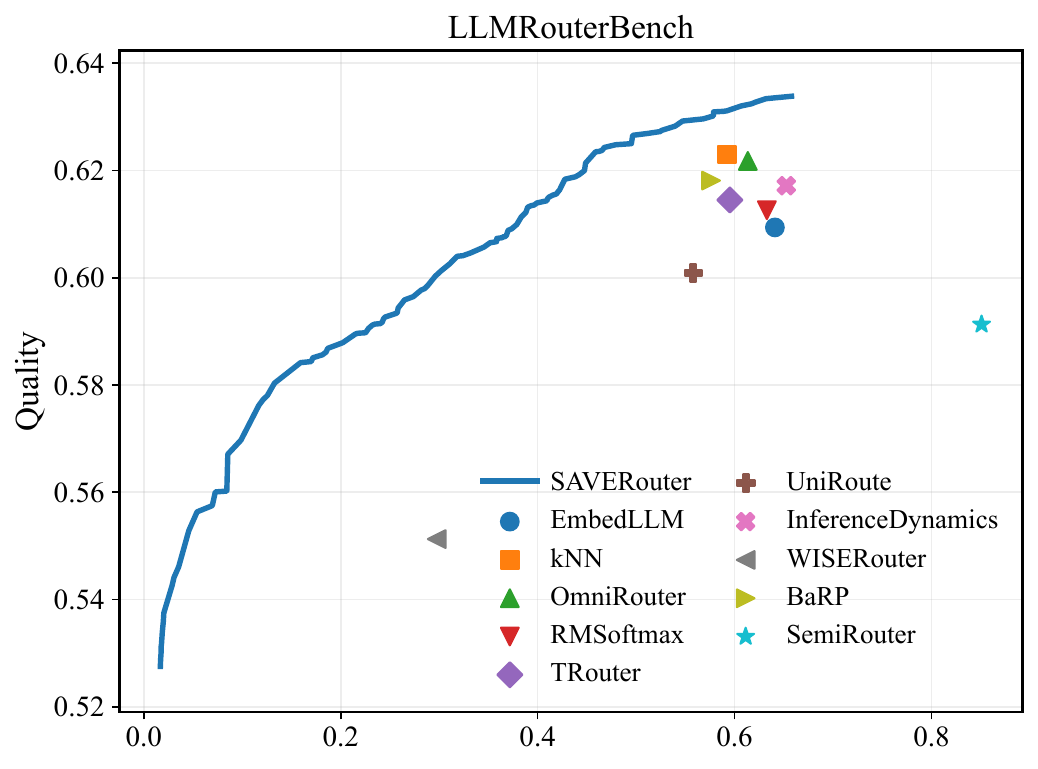}
        \caption{LLMRouterBench.}
    \end{subfigure}
    \hfill
    \begin{subfigure}[t]{0.49\textwidth}
        \centering
        \includegraphics[width=\linewidth]
        {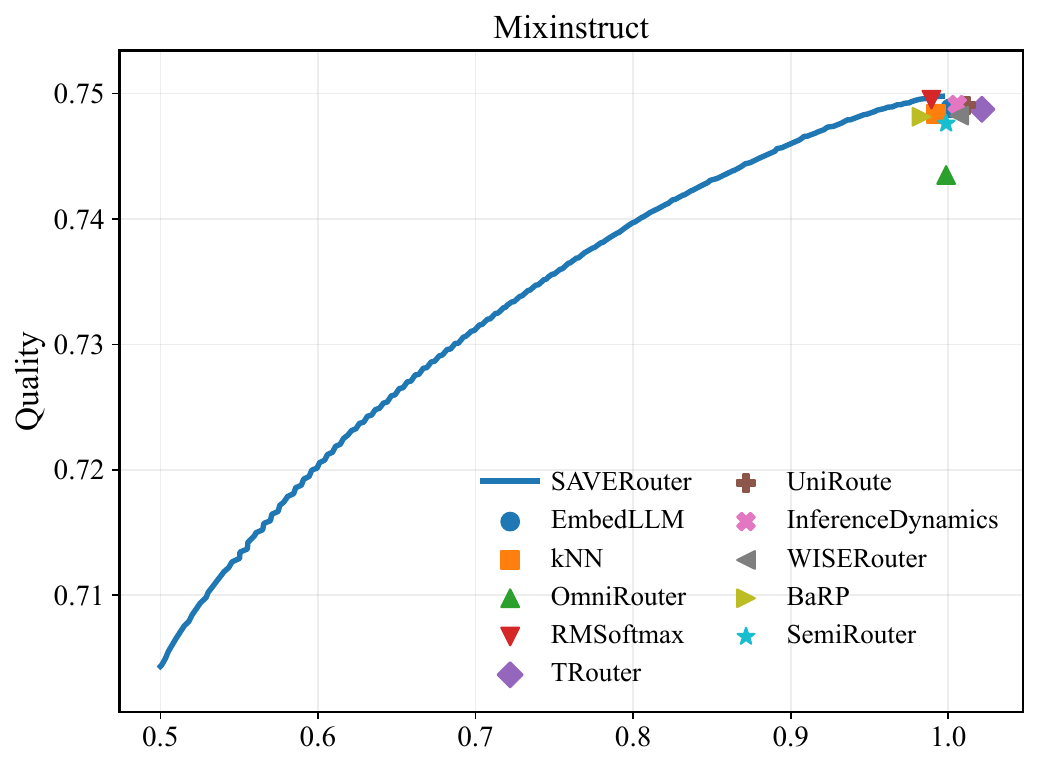}
        \caption{Mixinstruct.}
    \end{subfigure}

    \vspace{2mm}

    \begin{subfigure}[t]{0.49\textwidth}
        \centering
        \includegraphics[width=\linewidth]
        {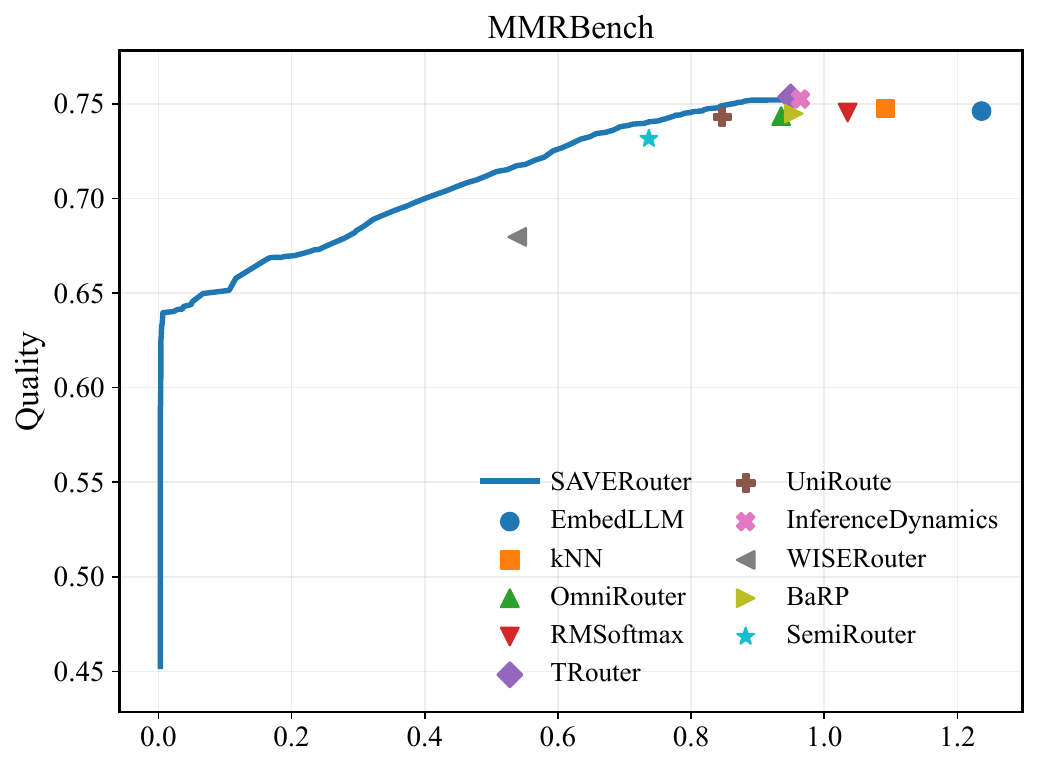}
        \caption{MMRBench.}
    \end{subfigure}
    \hfill
    \begin{subfigure}[t]{0.49\textwidth}
        \centering
        \includegraphics[width=\linewidth]
        {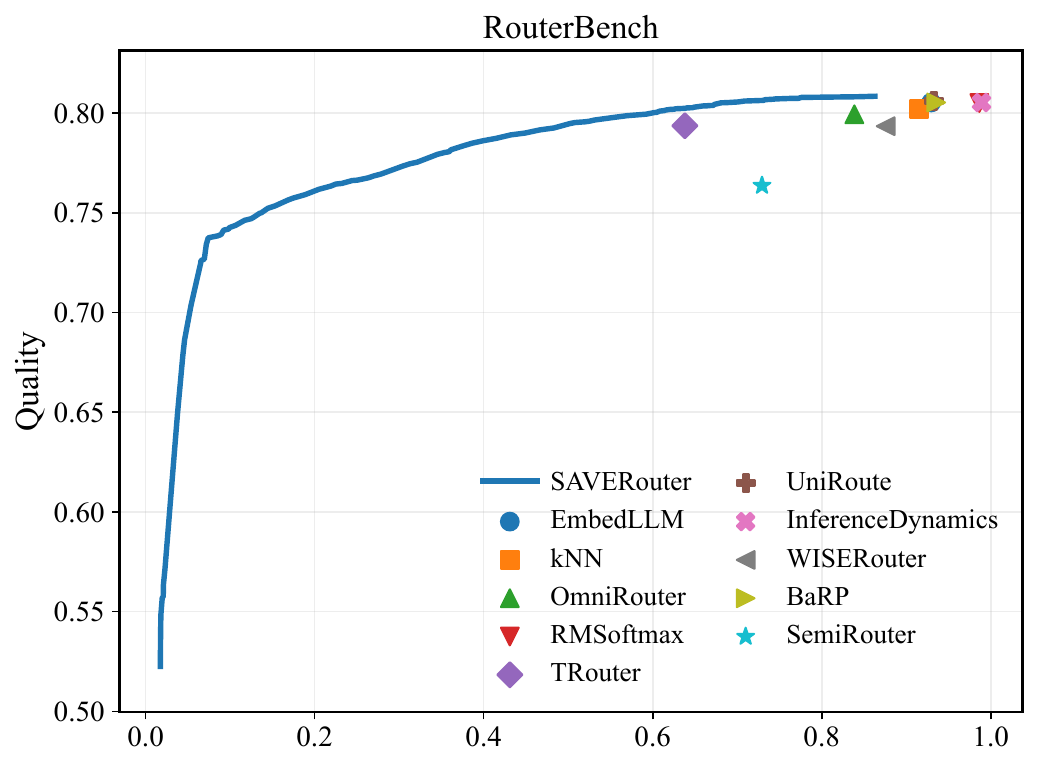}
        \caption{RouterBench.}
    \end{subfigure}

    \caption{
    \textbf{Full quality--cost frontiers across four routing benchmarks.}
    Each solid curve shows all evaluated \textsc{SaveRouter} operating points,
    while markers denote the peak-quality endpoint of each baseline method.
    The horizontal axis reports serving cost normalized by the best single
    model within each benchmark; lower is better, while higher quality is
    better.
    }
    \label{fig:full_pareto}
\end{figure*}

\subsection{Full Quality--Cost Frontiers}
\label{app:full_pareto}

Figure~\ref{fig:full_pareto} complements the scalar metrics in
Table~\ref{tab:main_results} by showing the full serving-time quality--cost
frontier of \textsc{SaveRouter} together with the peak-quality endpoint of
each baseline.

On LLMRouterBench, \textsc{SaveRouter} traces a broad frontier and reaches
$P_s=0.6338$, compared with 0.6230 for the strongest baseline peak endpoint.
Mixinstruct exhibits a different regime: baseline methods are tightly
concentrated near quality 0.75 and a cost ratio close to one, whereas
\textsc{SaveRouter} spans a substantially wider range of serving costs while
approaching the same saturated quality level.

On MMRBench, the strongest methods are more closely matched near the
high-quality end.
\textsc{SaveRouter} reaches $P_s=0.7540$, close to TRouter's 0.7536, while
providing additional lower-cost operating points.
On RouterBench, \textsc{SaveRouter} reaches $P_s=0.8084$ at a normalized
serving cost of approximately 0.86 at its peak-quality endpoint, whereas the
strongest baseline peak endpoint reaches 0.8062 at approximately 0.93.

Overall, the full frontiers show that \textsc{SaveRouter}'s performance is not
driven by a single favorable operating point.
Sparse supervision recovers a broad range of quality--cost trade-offs from
which different deployment preferences can be selected.
At the same time, the MMRBench results show that the frontier need not strictly
dominate every baseline at every cost level, motivating the use of both
frontier plots and scalar summary metrics.

\subsection{Performance across Quality Targets}
\label{app:quality_target_sensitivity}

Our main evaluation uses the quality $Q_b$ of the best single model as the
target operating point.
To examine how the results change under different quality requirements, we
evaluate each method at
\[
\rho Q_b,
\qquad
\rho\in\{0.90,0.925,0.95,0.975,1.00\}.
\]
For each target, we report CR, SA-BEP, and SA-CR@1M.
The $1.00Q_b$ column reproduces the operating points reported in
Table~\ref{tab:main_results}.

Tables~\ref{tab:target_llm}--\ref{tab:target_router} report each entry as
\[
\text{CR} \;/\; \text{SA-BEP} \;/\; \text{SA-CR@1M}.
\]
An entry of
$\infty/\infty/\infty$ indicates that the requested quality target is
unreachable.
When the target is reached but no positive per-query saving is achieved,
only SA-BEP and SA-CR are reported as $\infty$.

\begin{table*}[t]
\centering
\caption{
Performance across quality targets on LLMRouterBench.
Each entry reports CR / SA-BEP / SA-CR@1M; lower values are better.
}
\label{tab:target_llm}
\scriptsize
\setlength{\tabcolsep}{3.0pt}
\renewcommand{\arraystretch}{1.08}
\resizebox{\textwidth}{!}{
\begin{tabular}{lccccc}
\toprule
Method
& $0.90Q_b$
& $0.925Q_b$
& $0.95Q_b$
& $0.975Q_b$
& $1.00Q_b$ \\
\midrule

EmbedLLM
& 0.0286 / 7.7K / 0.0360
& 0.0953 / 8.2K / 0.1027
& 0.5246 / 15.6K / 0.5321
& 0.5246 / 15.6K / 0.5321
& 0.5246 / 15.6K / 0.5321 \\

kNN
& 0.0863 / 8.1K / 0.0938
& 0.1107 / 8.4K / 0.1182
& 0.1530 / 8.8K / 0.1605
& 0.2490 / 9.9K / 0.2564
& 0.3770 / 11.9K / 0.3845 \\

OmniRouter
& 0.1419 / 8.7K / 0.1493
& 0.1618 / 8.9K / 0.1693
& 0.2037 / 9.3K / 0.2112
& 0.2579 / 10.0K / 0.2653
& 0.3564 / 11.5K / 0.3638 \\

RMSoftmax
& 0.0285 / 7.7K / 0.0359
& 0.6328 / 20.2K / 0.6403
& 0.6328 / 20.2K / 0.6403
& 0.6328 / 20.2K / 0.6403
& 0.6328 / 20.2K / 0.6403 \\

TRouter
& 0.0328 / 7.7K / 0.0402
& 0.1500 / 8.7K / 0.1575
& 0.2447 / 9.8K / 0.2521
& 0.2995 / 10.6K / 0.3069
& 0.5221 / 15.6K / 0.5295 \\

UniRoute
& 0.0285 / 7.7K / 0.0359
& 0.1962 / 9.2K / 0.2036
& 0.5583 / 16.8K / 0.5657
& 0.5583 / 16.8K / 0.5657
& 0.5583 / 16.8K / 0.5657 \\

InferenceDyn.
& 0.0291 / 7.7K / 0.0366
& 0.0858 / 8.1K / 0.0932
& 0.1937 / 9.2K / 0.2011
& 0.4952 / 14.7K / 0.5026
& 0.4952 / 14.7K / 0.5026 \\

WISERouter
& 0.0831 / 711 / 0.0837
& 0.2636 / 885 / 0.2643
& $\infty$ / $\infty$ / $\infty$
& $\infty$ / $\infty$ / $\infty$
& $\infty$ / $\infty$ / $\infty$ \\

BaRP
& 0.1214 / 88.8K / 0.1994
& 0.1214 / 88.8K / 0.1994
& 0.1616 / 93.0K / 0.2396
& 0.2397 / 102.6K / 0.3177
& 0.3161 / 114.0K / 0.3941 \\

SemiRouter
& 0.4514 / 7.2K / 0.4553
& 0.4514 / 7.2K / 0.4553
& 0.4514 / 7.2K / 0.4553
& 0.8509 / 26.5K / 0.8548
& 0.8509 / 26.5K / 0.8548 \\

\specialrule{0.8pt}{2.5pt}{2pt}

\textbf{\textsc{SaveRouter}}
& \textbf{0.0182 / 4.1K / 0.0222}
& \textbf{0.0401 / 4.2K / 0.0441}
& \textbf{0.0853 / 4.4K / 0.0893}
& \textbf{0.1210 / 4.6K / 0.1250}
& \textbf{0.2320 / 5.3K / 0.2360} \\

\bottomrule
\end{tabular}
}
\end{table*}

\begin{table*}[t]
\centering
\caption{
Performance across quality targets on Mixinstruct.
Each entry reports CR / SA-BEP / SA-CR@1M; lower values are better.
}
\label{tab:target_mixinstruct}
\scriptsize
\setlength{\tabcolsep}{3.0pt}
\renewcommand{\arraystretch}{1.08}
\resizebox{\textwidth}{!}{
\begin{tabular}{lccccc}
\toprule
Method
& $0.90Q_b$
& $0.925Q_b$
& $0.95Q_b$
& $0.975Q_b$
& $1.00Q_b$ \\
\midrule

EmbedLLM
& 0.5000 / 432.7K / 0.7163
& 0.5000 / 432.7K / 0.7163
& 0.9924 / 28.60M / 1.2088
& 0.9924 / 28.60M / 1.2088
& 0.9924 / 28.60M / 1.2088 \\

kNN
& 0.5000 / 432.7K / 0.7163
& 0.5000 / 432.7K / 0.7163
& 0.8561 / 1.50M / 1.0725
& 0.8561 / 1.50M / 1.0725
& 0.9814 / 11.63M / 1.1977 \\

OmniRouter
& 0.5596 / 491.2K / 0.7759
& 0.5596 / 491.2K / 0.7759
& 0.6292 / 583.4K / 0.8455
& 0.8097 / 1.14M / 1.0261
& $\infty$ / $\infty$ / $\infty$ \\

RMSoftmax
& 0.5000 / 432.7K / 0.7163
& 0.5000 / 432.7K / 0.7163
& 0.9893 / 20.19M / 1.2056
& 0.9893 / 20.19M / 1.2056
& 0.9893 / 20.19M / 1.2056 \\

TRouter
& 0.5000 / 432.7K / 0.7163
& 0.5000 / 432.7K / 0.7163
& 0.6441 / 607.9K / 0.8604
& 0.9879 / 17.95M / 1.2043
& 1.0000 / $\infty$ / $\infty$ \\

UniRoute
& 0.5000 / 432.7K / 0.7163
& 0.5000 / 432.7K / 0.7163
& 1.0000 / $\infty$ / $\infty$
& 1.0000 / $\infty$ / $\infty$
& 1.0000 / $\infty$ / $\infty$ \\

InferenceDyn.
& 0.5000 / 432.7K / 0.7163
& 0.5000 / 432.7K / 0.7163
& 0.9996 / 565.46M / 1.2160
& 0.9996 / 565.46M / 1.2160
& 1.0000 / $\infty$ / $\infty$ \\

WISERouter
& 0.5000 / 36.1K / 0.5180
& 0.5000 / 36.1K / 0.5180
& 0.5402 / 39.2K / 0.5583
& 0.7172 / 63.8K / 0.7353
& 1.0000 / $\infty$ / $\infty$ \\

BaRP
& 0.5000 / 2.73M / 1.8634
& 0.5000 / 2.73M / 1.8634
& 0.5518 / 3.04M / 1.9152
& 0.7226 / 4.91M / 2.0860
& 0.9695 / 44.66M / 2.3329 \\

SemiRouter
& 0.5000 / 190.4K / 0.5952
& 0.5000 / 190.4K / 0.5952
& 0.9986 / 69.75M / 1.0938
& 0.9986 / 69.75M / 1.0938
& $\infty$ / $\infty$ / $\infty$ \\

\specialrule{0.8pt}{2.5pt}{2pt}

\textbf{\textsc{SaveRouter}}
& \textbf{0.5000 / 157.8K / 0.5789}
& \textbf{0.5000 / 157.8K / 0.5789}
& \textbf{0.5354 / 169.9K / 0.6143}
& \textbf{0.6770 / 244.3K / 0.7559}
& \textbf{0.9359 / 1.23M / 1.0148} \\

\bottomrule
\end{tabular}
}
\end{table*}

\begin{table*}[t]
\centering
\caption{
Performance across quality targets on MMRBench.
Each entry reports CR / SA-BEP / SA-CR@1M; lower values are better.
}
\label{tab:target_mmr}
\scriptsize
\setlength{\tabcolsep}{3.0pt}
\renewcommand{\arraystretch}{1.08}
\resizebox{\textwidth}{!}{
\begin{tabular}{lccccc}
\toprule
Method
& $0.90Q_b$
& $0.925Q_b$
& $0.95Q_b$
& $0.975Q_b$
& $1.00Q_b$ \\
\midrule

EmbedLLM
& 1.0000 / $\infty$ / $\infty$
& 1.0000 / $\infty$ / $\infty$
& 1.0000 / $\infty$ / $\infty$
& 1.0000 / $\infty$ / $\infty$
& 1.0000 / $\infty$ / $\infty$ \\

kNN
& 0.5503 / 12.2K / 0.5557
& 0.7515 / 22.0K / 0.7569
& 0.9274 / 75.3K / 0.9329
& 1.0000 / $\infty$ / $\infty$
& 1.0000 / $\infty$ / $\infty$ \\

OmniRouter
& 0.2911 / 7.7K / 0.2965
& 0.4700 / 10.3K / 0.4755
& 0.6581 / 16.0K / 0.6635
& 0.8215 / 30.6K / 0.8269
& $\infty$ / $\infty$ / $\infty$ \\

RMSoftmax
& 1.0350 / $\infty$ / $\infty$
& 1.0350 / $\infty$ / $\infty$
& 1.0350 / $\infty$ / $\infty$
& 1.0350 / $\infty$ / $\infty$
& 1.0350 / $\infty$ / $\infty$ \\

TRouter
& 0.2103 / 6.9K / 0.2158
& 0.6730 / 16.7K / 0.6784
& 0.6730 / 16.7K / 0.6784
& 0.6730 / 16.7K / 0.6784
& 0.8396 / 34.1K / 0.8451 \\

UniRoute
& 0.8173 / 29.9K / 0.8227
& 0.8173 / 29.9K / 0.8227
& 0.8173 / 29.9K / 0.8227
& 0.8173 / 29.9K / 0.8227
& $\infty$ / $\infty$ / $\infty$ \\

InferenceDyn.
& 0.9641 / 152.4K / 0.9696
& 0.9641 / 152.4K / 0.9696
& 0.9641 / 152.4K / 0.9696
& 0.9641 / 152.4K / 0.9696
& 0.9641 / 152.4K / 0.9696 \\

WISERouter
& 0.5180 / 1.2K / 0.5186
& $\infty$ / $\infty$ / $\infty$
& $\infty$ / $\infty$ / $\infty$
& $\infty$ / $\infty$ / $\infty$
& $\infty$ / $\infty$ / $\infty$ \\

BaRP
& 0.2468 / 80.7K / 0.3076
& 0.3553 / 94.2K / 0.4161
& 0.5298 / 129.2K / 0.5905
& 0.6804 / 190.0K / 0.7411
& 0.9052 / 640.8K / 0.9660 \\

SemiRouter
& 0.7366 / 12.5K / 0.7399
& 0.7366 / 12.5K / 0.7399
& 0.7366 / 12.5K / 0.7399
& 0.7366 / 12.5K / 0.7399
& $\infty$ / $\infty$ / $\infty$ \\

\specialrule{0.8pt}{2.5pt}{2pt}

\textbf{\textsc{SaveRouter}}
& \textbf{0.1909 / 5.1K / 0.1950}
& \textbf{0.3225 / 6.0K / 0.3266}
& \textbf{0.4633 / 7.6K / 0.4674}
& \textbf{0.5929 / 10.0K / 0.5970}
& \textbf{0.7765 / 18.2K / 0.7806} \\

\bottomrule
\end{tabular}
}
\end{table*}

\begin{table*}[t]
\centering
\caption{
Performance across quality targets on RouterBench.
Each entry reports CR / SA-BEP / SA-CR@1M; lower values are better.
}
\label{tab:target_router}
\scriptsize
\setlength{\tabcolsep}{3.0pt}
\renewcommand{\arraystretch}{1.08}
\resizebox{\textwidth}{!}{
\begin{tabular}{lccccc}
\toprule
Method
& $0.90Q_b$
& $0.925Q_b$
& $0.95Q_b$
& $0.975Q_b$
& $1.00Q_b$ \\
\midrule

EmbedLLM
& 0.0742 / 24.8K / 0.0972
& 0.2648 / 31.3K / 0.2878
& 0.9296 / 326.4K / 0.9526
& 0.9296 / 326.4K / 0.9526
& 0.9296 / 326.4K / 0.9526 \\

kNN
& 0.0655 / 24.6K / 0.0885
& 0.1443 / 26.9K / 0.1672
& 0.4101 / 39.0K / 0.4331
& 0.7455 / 90.3K / 0.7684
& $\infty$ / $\infty$ / $\infty$ \\

OmniRouter
& 0.1617 / 27.4K / 0.1847
& 0.1811 / 28.1K / 0.2040
& 0.3522 / 35.5K / 0.3752
& 0.6616 / 67.9K / 0.6846
& $\infty$ / $\infty$ / $\infty$ \\

RMSoftmax
& 0.9865 / 1.70M / 1.0095
& 0.9865 / 1.70M / 1.0095
& 0.9865 / 1.70M / 1.0095
& 0.9865 / 1.70M / 1.0095
& 0.9865 / 1.70M / 1.0095 \\

TRouter
& 0.0901 / 25.3K / 0.1131
& 0.3151 / 33.5K / 0.3381
& 0.4450 / 41.4K / 0.4680
& 0.5987 / 57.3K / 0.6217
& $\infty$ / $\infty$ / $\infty$ \\

UniRoute
& 0.0741 / 24.8K / 0.0970
& 0.9327 / 341.3K / 0.9557
& 0.9327 / 341.3K / 0.9557
& 0.9327 / 341.3K / 0.9557
& 0.9327 / 341.3K / 0.9557 \\

InferenceDyn.
& 0.0737 / 24.8K / 0.0966
& 0.2626 / 31.2K / 0.2856
& 0.9892 / 2.13M / 1.0122
& 0.9892 / 2.13M / 1.0122
& 0.9892 / 2.13M / 1.0122 \\

WISERouter
& 0.0742 / 2.3K / 0.0763
& 0.1901 / 2.6K / 0.1922
& 0.3779 / 3.4K / 0.3800
& 0.7800 / 9.5K / 0.7821
& $\infty$ / $\infty$ / $\infty$ \\

BaRP
& 0.0848 / 228.8K / 0.2942
& 0.1143 / 236.4K / 0.3237
& 0.2143 / 266.5K / 0.4237
& 0.4710 / 395.8K / 0.6804
& 0.9355 / 3.25M / 1.1449 \\

SemiRouter
& 0.7294 / 43.6K / 0.7412
& 0.7294 / 43.6K / 0.7412
& $\infty$ / $\infty$ / $\infty$
& $\infty$ / $\infty$ / $\infty$
& $\infty$ / $\infty$ / $\infty$ \\

\specialrule{0.8pt}{2.5pt}{2pt}

\textbf{\textsc{SaveRouter}}
& \textbf{0.0652 / 14.6K / 0.0788}
& \textbf{0.1108 / 15.3K / 0.1244}
& \textbf{0.2251 / 17.6K / 0.2388}
& \textbf{0.3862 / 22.2K / 0.3999}
& \textbf{0.6810 / 42.6K / 0.6946} \\

\bottomrule
\end{tabular}
}
\end{table*}

The target sweep reveals two complementary effects.
Relaxing the quality requirement generally reduces serving cost and shortens
payback, but it can also change which method reaches break-even earliest.
As the target approaches $Q_b$, the ability to preserve high-quality routing
becomes increasingly important.

On LLMRouterBench, \textsc{SaveRouter} achieves the lowest CR and SA-CR@1M
at every evaluated target.
Its CR increases from 0.0182 at $0.90Q_b$ to 0.2320 at $Q_b$, while SA-BEP
remains between 4.1K and 5.3K queries.
WISERouter reaches break-even earlier at the two most relaxed targets because
of its very small supervision expenditure, but cannot reach $0.95Q_b$ or
higher.
Thus, very early payback at a relaxed target does not necessarily imply access
to the high-quality region of the routing frontier.

Mixinstruct exhibits the strongest saturation effect.
At $0.90Q_b$ and $0.925Q_b$, many methods reach the same minimum-cost
operating point with CR=0.5, making supervision expenditure the dominant
difference in SA-BEP and SA-CR.
As the target becomes stricter, however, several methods move close to
CR=1 or lose positive serving-time savings, whereas \textsc{SaveRouter}
continues to reach the full target with CR=0.9359.
The distinction between methods therefore becomes substantially larger near
the strict quality target despite highly saturated peak quality.

On MMRBench, \textsc{SaveRouter} achieves the lowest CR and SA-CR@1M across
all five targets and retains a finite break-even point throughout the sweep.
WISERouter pays back earlier at $0.90Q_b$ but cannot reach the higher targets.
As the quality requirement increases, several conventional routers approach
or exceed the serving cost of the best single model, while
\textsc{SaveRouter} maintains a lower-cost target-quality operating point.

RouterBench further illustrates why serving-time cost and supervision
expenditure should be considered jointly.
WISERouter achieves the earliest payback at relaxed targets but cannot reach
the full $Q_b$ target.
At $0.95Q_b$, BaRP obtains a slightly lower CR than \textsc{SaveRouter}
(0.2143 versus 0.2251), yet its larger supervision expenditure leads to a
substantially higher SA-CR@1M (0.4237 versus 0.2388).
A lower serving-time CR therefore does not necessarily imply lower
supervision-amortized cost.

Overall, the conclusions are not specific to the single $Q_b$ operating
point.
Relaxed targets can favor methods with very low supervision expenditure,
whereas stricter targets increasingly reward methods that preserve a broad
high-quality routing frontier.
Reporting CR together with SA-BEP and SA-CR therefore exposes trade-offs that
are hidden by either serving-time efficiency or payback alone.

\subsection{Sensitivity to Feedback Caching for BaRP}
\label{app:barp_cache}

Our main evaluation adopts fresh-feedback accounting, under which each
bandit interaction is treated as a separate model invocation, including
repeated requests for the same query--model pair.
We additionally consider cached-feedback accounting, where only the first
interaction with a query--model pair incurs its model-execution cost and
subsequent interactions reuse the previously acquired feedback at no
additional supervision cost.

This analysis changes only the supervision-cost accounting for BaRP.
Its routing predictions and serving-time quality--cost frontier are held
fixed, so $P_s$ and CR remain unchanged.
Table~\ref{tab:barp_cache} reports the resulting SA-BEP and SA-CR values.

\begin{table}[t]
\centering
\caption{
Sensitivity of BaRP to fresh versus cached feedback accounting.
$P_s$ and CR are unchanged because caching affects only supervision
expenditure.
Lower SA-BEP and SA-CR@1M are better.
}
\label{tab:barp_cache}

\small
\setlength{\tabcolsep}{4pt}
\begin{tabular}{lcccccc}
\toprule
& $P_s$ $\uparrow$
& CR $\downarrow$
& \multicolumn{2}{c}{SA-BEP $\downarrow$}
& \multicolumn{2}{c}{SA-CR@1M $\downarrow$} \\
\cmidrule(lr){4-5}
\cmidrule(lr){6-7}
Benchmark
& &
& Cached & Fresh
& Cached & Fresh \\
\midrule

LLMRouterBench
& 0.6181
& 0.3161
& 10.8K
& 114.0K
& 0.3235
& 0.3941 \\

Mixinstruct
& 0.7482
& 0.9695
& 6.13M
& 44.66M
& 1.1565
& 2.3329 \\

MMRBench
& 0.7448
& 0.9052
& 57.6K
& 640.8K
& 0.9107
& 0.9660 \\

RouterBench
& 0.8053
& 0.9355
& 329.2K
& 3.25M
& 0.9567
& 1.1449 \\

\bottomrule
\end{tabular}
\end{table}

Caching substantially reduces BaRP's accounted supervision expenditure.
Across the four benchmarks, SA-BEP is reduced by a factor of approximately
$7.3$--$11.1$ relative to fresh-feedback accounting.
The largest change in SA-CR occurs on Mixinstruct, where SA-CR@1M decreases
from 2.3329 to 1.1565.
On RouterBench, caching reduces SA-CR@1M from 1.1449 to 0.9567, moving the
one-million-query horizon from before to after break-even.

These results show that BaRP's supervision-amortized cost is sensitive to
whether repeated interactions require fresh model executions.
Our main results correspond to the fresh-feedback setting, while this
sensitivity analysis quantifies the potential benefit when previously
acquired query--model feedback can be cached and reused.

\section{Sparse Model-Pool Expansion}
\label{sec:model-pool-expansion}

We further study whether \textsc{SaveRouter} can incorporate a newly available
model without exhaustively evaluating it on the training set.
For each benchmark, we remove one model from the original pool and treat it as
an arriving model $m^+$.
We consider low-, median-, and high-cost arrivals according to their average
training-time serving cost.

The existing model pool uses the same $K=4$ sparse supervision as in the main
experiments.
To incorporate $m^+$, we acquire its feedback on approximately 10\% of the
training queries, sampled within query groups, and retrain \textsc{SaveRouter}
using the combined observations.
We compare this setting, denoted \textsc{SaveRouter}-10\%, with
\textsc{SaveRouter}-100\%, which observes the arriving model on all available
training queries, and with fully supervised dense routers over the expanded
model pool.
All methods route over the same expanded candidate pool at test time.

\begin{table*}[t]
\centering
\caption{
Sparse model-pool expansion with $K=4$ supervision for the existing model
pool.
Subscripts 10 and 100 denote using approximately 10\% and 100\% of the
available training feedback for the arriving model.
$\rho_{C,+}$ is the acquisition cost of the 10\% feedback relative to fully
labeling the arriving model.
Dense $P_s$ and CR are independent per-metric optima over the fully supervised
baselines and need not correspond to the same router.
Higher $P_s$ is better; lower $\rho_{C,+}$, CR, SA-BEP, and SA-CR@1M are
better.
}
\label{tab:low-cost-pool-expansion}

\resizebox{\textwidth}{!}{
\begin{tabular}{llrrrrrrrrrr}
\toprule
Benchmark
& Arrival
& $|\Omega^+|$
& $\rho_{C,+}$
& $P_{s,10}$
& $P_{s,100}$
& $P_{s,\mathrm{dense}}$
& $\mathrm{CR}_{10}$
& $\mathrm{CR}_{100}$
& $\mathrm{CR}_{\mathrm{dense}}$
& SA-BEP$_{10}$
& SA-CR@1M$_{10}$ \\
\midrule

LLMRouterBench
& Cheap
& 254
& 11.92\%
& 0.6276
& 0.6273
& 0.6230
& 0.2296
& 0.2261
& 0.3564
& 4,921
& 0.2334 \\

& Median
& 254
& 9.35\%
& 0.6263
& 0.6340
& 0.6230
& 0.2650
& 0.2287
& 0.3564
& 5,610
& 0.2691 \\

& Expensive
& 254
& 10.97\%
& 0.6339
& 0.6305
& 0.6230
& 0.2606
& 0.2412
& 0.3564
& 3,887
& 0.2635 \\

\midrule

Mixinstruct
& Cheap
& 2,200
& 10.00\%
& 0.7486
& 0.7496
& 0.7495
& 0.9847
& 0.9516
& 0.9814
& 6,088,547
& 1.0779 \\

& Median
& 2,200
& 10.00\%
& 0.7498
& 0.7498
& 0.7495
& 0.9354
& 0.9354
& 0.9814
& 1,257,581
& 1.0166 \\

& Expensive
& 2,200
& 10.00\%
& 0.7498
& 0.7497
& 0.7495
& 0.9368
& 0.9569
& 0.9814
& 1,287,017
& 1.0181 \\

\midrule

MMRBench
& Cheap
& 211
& 9.45\%
& 0.7495
& 0.7482
& 0.7536
& 0.7959
& 0.7976
& 0.8396
& 13,195
& 0.7986 \\

& Median
& 211
& 9.50\%
& 0.7467
& 0.7477
& 0.7536
& 0.7475
& 0.7212
& 0.8396
& 10,597
& 0.7502 \\

& Expensive
& 211
& 10.35\%
& 0.7471
& 0.7458
& 0.7536
& 0.9752
& 0.8875
& 0.8396
& 80,625
& 0.9772 \\

\midrule

RouterBench
& Cheap
& 771
& 10.47\%
& 0.8091
& 0.8091
& 0.8062
& 0.6458
& 0.6409
& 0.9296
& 54,742
& 0.6652 \\

& Median
& 771
& 10.48\%
& 0.8080
& 0.8092
& 0.8062
& 0.6584
& 0.6408
& 0.9296
& 56,095
& 0.6775 \\

& Expensive
& 771
& 10.48\%
& 0.8074
& 0.8090
& 0.8062
& 0.6651
& 0.6381
& 0.9296
& 40,078
& 0.6785 \\

\bottomrule
\end{tabular}
}
\end{table*}

Across the 12 arrival cases, sparse onboarding requires only 211--2,200
new-model observations.
The associated acquisition cost is 9.35\%--11.92\% of fully evaluating the
arriving model, eliminating roughly 90\% of its supervision expenditure.

Despite this reduction, \textsc{SaveRouter}-10\% remains competitive with both
full new-model supervision and dense routing.
Relative to the best fully supervised dense baselines,
\textsc{SaveRouter}-10\% achieves a higher $P_s$ in 8 of the 12 arrival
cases and a lower CR in 10 of 12 cases.
On LLMRouterBench and RouterBench, it achieves both higher peak quality and
lower target-quality serving cost than the dense optima for all three arrival
types.

The comparison with \textsc{SaveRouter}-100\% further shows that exhaustive
feedback from the arriving model is not consistently necessary.
On several arrivals, increasing new-model supervision improves one operating
metric but leaves another nearly unchanged or worse.
For example, the 10\% setting is already nearly indistinguishable from full
new-model supervision on the median- and high-cost Mixinstruct arrivals, while
on RouterBench the quality gap remains very small across all three cases.
This indicates that much of the useful information about an arriving model can
often be recovered from a relatively small number of strategically distributed
observations.

The behavior is not uniform across benchmarks.
On MMRBench, the fully supervised dense baseline retains higher $P_s$ for all
three arrivals, and the expensive arrival yields a relatively high CR of
0.9752 under sparse onboarding.
Mixinstruct also remains difficult from a payback perspective:
SA-CR@1M stays above one for all three arrivals despite strong routing quality.
These cases indicate that sparse onboarding is most effective when limited
feedback is sufficient to characterize the new model's relative capability and
cost within the existing pool.

Overall, the experiment suggests that expanding the candidate model pool need
not require fully evaluating every newly available model on the training set.
Across most arrival scenarios, approximately 10\% new-model feedback preserves
competitive routing quality while avoiding the majority of the additional
supervision expenditure.
This experiment studies retrained sparse onboarding rather than zero-shot or
parameter-preserving online admission.

\section{Broader Connections and Discussion}
\label{sec:discussion}

\noindent\textbf{Connection to efficient benchmarking.}
\textsc{SaveRouter} is related to efficient benchmarking, which reduces the
number of model--example evaluations needed to estimate benchmark outcomes
\citep{perlitz2024efficient,li2024active,zhang2026efficient}.
Both exploit redundancy in the evaluation matrix, but their objectives differ:
benchmarking aims to preserve aggregate quantities such as model scores or
rankings, whereas \textsc{SaveRouter} uses sparse observations to support
query-level routing decisions.
Thus, accurate recovery of the full matrix is unnecessary as long as the
observed feedback is sufficient to preserve the downstream decision.

\noindent\textbf{Connection to active testing and active learning.}
Our acquisition problem also resembles active testing and active learning,
which allocate limited evaluation or labeling effort to informative examples
\citep{kossen2021active}.
The key difference is that \textsc{SaveRouter} acquires
\emph{query--model pairs} rather than query labels alone:
for the same query, evaluating different models can provide different amounts
of routing-relevant information.
The acquisition problem therefore couples query structure with heterogeneous
model capabilities and is optimized for the eventual routing decision rather
than full outcome reconstruction.

\noindent\textbf{Connection to cold-start model selection.}
Sparse supervision is also relevant when the candidate model pool changes over
time.
A newly available model has initially unknown performance on the historical
workload, making exhaustive evaluation potentially expensive.
Our model-pool expansion results show that a small fraction of new-model
feedback can often be sufficient to incorporate it into an existing routing
system.
This connects model-pool expansion to cold-start and experimental-design
settings, where limited measurements are used to characterize a new
alternative before deployment.

Overall, these connections highlight a common principle:
when performance feedback is costly and the final objective is a decision,
the goal need not be complete recovery of the underlying observation matrix.
Instead, supervision can be selected according to its value for the downstream
decision and the extent to which its acquisition cost can be recovered through
subsequent serving-time savings.

\end{document}